\documentclass[print]{elsarticle}

\usepackage{lineno,hyperref}

\usepackage{graphicx}
\usepackage{color}
\usepackage{amssymb}
\usepackage{subfigure}
\usepackage{amsmath}
\newtheorem{definition}{Definition}[section]
\newtheorem{theorem}{Theorem}[section]

\usepackage{bm}
\usepackage[misc]{ifsym}
\usepackage{appendix}

\begin{document}

\begin{frontmatter}

\title{Efficient Use of heuristics for accelerating XCS-based Policy Learning in Markov Games}

		\author[mymainaddress]{Hao Chen}
		\ead{nudtchenhao15a@163.com}
		\author[mymainaddress]{Chang Wang}
		\author[mymainaddress]{Jian Huang\corref{mycorrespondingauthor}}
		\cortext[mycorrespondingauthor]{Corresponding author}
		\ead{nudtjhuang@hotmail.com}
		
		\ead{wangchang07@nudt.edu.cn}
\author[mymainaddress]{Jianxing Gong}
\ead{fj\_gjx@qq.com}

\address[mymainaddress]{College of Intelligence Science and Technology, National University of Defense Technology, Changsha, Hunan, 410073, China}

\begin{abstract}
In Markov games, playing against non-stationary opponents with learning ability is still challenging for reinforcement learning (RL) agents, because the opponents can evolve their policies concurrently. This increases the complexity of the learning task and slows down the learning speed of the RL agents.
This paper proposes efficient use of rough heuristics to speed up policy learning when playing against concurrent learners. Specifically, we propose an algorithm that can efficiently learn explainable and generalized action selection rules by taking advantages of the representation of quantitative heuristics and an opponent model with an eXtended classifier system (XCS) in zero-sum Markov games. A neural network is used to model the opponent from their behaviors and the corresponding policy is inferred for action selection and rule evolution.
In cases of multiple heuristic policies, we introduce the concept of Pareto optimality for action selection. Besides, taking advantages of the condition representation and matching mechanism of XCS, the heuristic policies and the opponent model can provide guidance for situations with similar feature representation. 
Furthermore, we introduce an accuracy-based eligibility trace mechanism to speed up rule evolution, i.e., classifiers that can match the historical traces are reinforced according to their accuracy. 
We demonstrate the advantages of the proposed algorithm over several benchmark algorithms in a soccer and a thief-and-hunter scenarios.


\end{abstract}


\end{frontmatter}


\section{Introduction}
\label{intro}
In a single-agent system (SAS), the environment is stationary and a reinforcement learning (RL) task is usually modeled as a Markov decision process (MDP) \cite{sutton2018reinforcement}. Compared with SAS, learning in a multi-agent system (MAS) \cite{bucsoniu2010multi} is more difficult because an agent not only interacts with the environment but also with the other agents in the scenario, e.g., in Markov games \cite{littman1994markov}. If every other player in a Markov game follows stationary policies, the learning task can also be modeled as an MDP \cite{bowling2002multiagent} where the traditional RL algorithms for SAS can be used by the RL agent. However, in many Markov games, the opponents may exhibit more sophisticated behaviors, such as learning concurrently with the RL agent.  In this case, the environment is non-stationary and the traditional MDP framework is no longer suitable. 


Trying to solve zero-sum Markov games with a RL opponent, Littman et al. \cite{littman1994markov} have proposed the Minimax-Q learning algorithm, which is an extension of Q-learning \cite{watkins1992q} in SAS. Similar to Q-learning, Minimax-Q may be space-inefficient and time-consuming when handling large problems. To alleviating this problem and speed up the learning process, a variety of accelerating techniques and corresponding Q-table based RL (QbRL) algorithms have been proposed. For example, Minimax-QS \cite{Ribeiro2002Experience}, Minimax-Q($ \lambda $) \cite{Banerjee2001Fast}, and heuristically-accelerated Minimax-Q (HAMMQ) \cite{Bianchi2007Heuristic, Bianchi2014Heuristically} have been proposed based on spatial, temporal, and action generalization, respectively. Although the learned Q-tables are easy to understand, the application of these QbRL algorithms have limitations for the huge state-action space, which makes the optimal policy search process difficult. Besides, these algorithms assume the opponent follows an optimal policy based on Nash equilibrium and therefore makes decisions conservatively.

One efficient way of handling large state-action space is to leverage the idea of generalization, by which a good approximate solution can be found with limited computational resources \cite{sutton2018reinforcement}.
Neural networks have been used as function approximators to generalize from obtained samples of the state-action value function. Lowe et al. \cite{DBLP:journals/corr/LoweWTHAM17} have proposed (multi-agent deep deterministic policy gradient) MADDPG for multi-agent cooperation and competition by extending deep deterministic policy gradient (DDPG) \cite{lillicrap2016continuous} to MAS. He et al. \cite{he2016opponent} have proposed (deep reinforcement learning opponent network) DRON which incorporates opponents' behaviors into deep Q-network (DQN) \cite{mnih2015human}. However, the learning results of these neural-network-based RL (NNbRL) algorithms lack explainability for humans to understand why an action is selected in case of a given state.

The eXtended classifier system (XCS) \cite{wilson1995classifier, wilson1998generalization} is an accuracy based learning classifier system (LCS) \cite{Holland1992Adaptation} used for SAS. In XCS, the RL part is responsible for payoff prediction and the genetic algorithms (GA) part is responsible for evolving the population of action selection rules. 
Compared with QbRL algorithms with a Q-table, XCS can learn generalized action selection rules, thus greatly reducing the memory space required for policy storage. Moreover, compared with NNbRL algorithms, the learned action selection rules are readable and easy to understand.
In the literature, XCS has been used in both cooperative RL tasks \cite{li2006refined,chen2009hierarchical} and competitive RL tasks \cite{inoue2005exploring, lode2010adaption}. However, these algorithms treat the agents independently without considering their interdependencies.
In this paper, we leverage the advantages of XCS and extend them to Markov games.

Another efficient approach to boost the learning process is to use the heuristic knowledge efficiently. A feasible way is to reuse the learned policies, such as RL-CD \cite{Silva2006Dealing}, Bayes-Pepper \cite{hernandezleal2017towards}, and Deep Bayesian policy reuse+ (Deep BPR+) \cite{NIPS2018_7374}. These algorithms maintain a policy library that records previously learned policies. In the application phase, they detect the policies of other participants during online interactions and then reuse the optimal response policy or continue learning on this basis. Another way is learning from scratch and then speed up the learning by selecting a suitable heuristic policy for explorations, such as the $\pi$-reuse exploration strategy \cite{Veloso06probabilisticpolicy}, the HAMMQ \cite{Bianchi2007Heuristic, Bianchi2014Heuristically} and $\pi$-selection algorithm \cite{piselection}. In these algorithms, the heuristics directly guide the action selection of the agent in specific states it covers.
However, the provided heuristics are often suboptimal and only cover parts of the whole state space \cite{zhang2020kogun}, thus the uncovered states are ignored by the heuristics. As a result, it is critical to make efficient use of the suboptimal prior and carry out policy learning on this basis. In this paper, we use the heuristic policies by quantifying the heuristic policies into the classifier representation of XCS. In this way, the quantified heuristics  in the classifier can provide guidance for similar states that can match the condition of the classifier.

Other players' models are also useful for the learning agents in MAS. Specifically, leveraging other players' models, an agent can predict the behaviors of the others and makes the best responses accordingly. A natural idea is to observe other players' action frequencies and use these to predict their behaviors in specific states, such as JAL \cite{claus1998dynamics} and non-stationary converging policies (NSCP) \cite{weinberg2004best}. Deviation-based best response (DBBR) \cite{Ganzfried2011Game} builds the opponent model by combining the opponent's action frequencies and the precomputed equilibrium strategy. One limitation of these methods is that the learned models are state specific without generalization, i.e., they can only be used in previously accessed states. In contrast, we build an explicit opponent model using a neural network and update it with the opponent's behavior sequences. Benefiting from the condition representation mechanism of XCS, the opponent model can predict the opponent's action in similar states. Similar ideas can also be found in NNbRL algorithms, such as deep reinforcement opponent network (DRON) \cite{he2016opponent} and deep inference Q-network (DPIQN) \cite{DPIQN}.
One difference between these algorithms and our work is that we do not learn a mixture of experts architecture or hidden features of the opponent's policy. Instead, we model the opponent explicitly and use the opponent model for action selection and the associated rules evolution.

In a recent paper, we have proposed the use of XCS in zero-sum Markov games to learn general, accurate and interpretable action selection rules \cite{XOMQ}. In this paper, we extend our previous work and investigate the efficient use of heuristics to further improve the learning performance along with the evolution of XCS classifiers.
We propose a novel XCS-based algorithm named heuristic accelerated Markov games with XCS (HAMXCS) to solve competitive Markov games, which incorporates the provided rough heuristic policies and an opponent model for best response policy learning. In our settings, the players have full observability of the environment, but no prior knowledge about others' goals.
The main contributions of the paper are as follows:

\begin{enumerate}
	\item [$ \bullet $] We introduce the quantitative representation of heuristics integrated with XCS classifiers. As a result, the XCS classifiers can be used for action selection in similar states, and also updated according to the heuristic policies.
	\item [$\bullet $] We construct an opponent model with a neural network and integrate it into the proposed algorithm for predicting the opponent's behaviors, while the prediction is simultaneously used to select actions and evolve classifiers. Combing with the generalization capability of the neural network and the representation mechanism of XCS classifiers, the opponent model can be used to predict the opponent's behavior in similar states.
	\item[$ \bullet $] We propose an accuracy-based eligibility trace mechanism to accelerate policy learning. Using the fitness parameter of XCS, classifiers that match the historical traces are updated accounting for their accuracy.
\end{enumerate}

The rest of the paper is organized as follows. Section \ref{related_works} introduces the related work of XCS and policy learning methods in Markov games. Section \ref{HAMXCS} presents the proposed HAMXCS algorithm in detail. Section \ref{experiment} describes the experiments and discusses the results. Finally, Section \ref{conclusion} concludes the paper.

\section{Related work}
\label{related_works}

\subsection{Markov games} \label{pl_in_mg}

Markov game \cite{littman1994markov} is a framework which was proposed for policy learning in MAS. Formally, a Markov Game $ \mathcal{M} $ involving $ n $ players can be presented as a tuple $ \mathcal{M}= \left\langle n, \mathcal{S}, \mathcal{A}_{1}, \dots,\mathcal{A}_{n}, \mathcal{P}, r_{1}, \dots, r_{n} \right\rangle $. The environment occupies states $ s \in \mathcal{S} $, in which, each player selects action $ a_i \in \mathcal{A}_i $ at each time step. $ \mathcal{P} $ is a state transition function, $ \mathcal{P}: \mathcal{S} $ $ \times $ $ \mathcal{A}_{1} \dots$ $ \times $ $\mathcal{A}_{n} $ $ \mapsto $ $ \Pi\left( S \right) $, which describes the probability of changing from the current state $ s $ to the next state $ s' $ when the players execute $ a_{1}, \dots, a_{n} $, respectively. Each player has a reward function $ r_{i}: $ $ \mathcal{S} $ $ \times $ $ \mathcal{A}_{1} \dots$ $ \times $ $\mathcal{A}_{n} $ $ \mapsto $ $ \mathbb{R} $ and tries to maximize its own total expected return $ R_i = \sum_{t=0}^{T} \gamma^t r_i^t$, where $\gamma \in \left[ 0,1\right) $ is the discount factor and $ T $ is the time horizon.

In this paper, we focus on zero-sum Markov games, in which a RL agent plays against an opponent. Denote $ \mathcal{A}$ for the agent's action set and $  \mathcal{O}$ for the opponent's. 

\subsection{QbRL for Markov games}

In the framework of Markov games, Littman has proposed the Minimax-Q learning algorithm \cite{littman1994markov} that combines the Minimax algorithm in games and Q-learning \cite{watkins1992q} in RL. Minimax-Q is similar to Q-learning except that the $ \mathit{max} $ operator is replaced by the $ \mathit{minimax} $ operator of Q-learning. The update rule of Minimax-Q is as follows:
\begin{equation}\label{euq1}
Q\left( {s,a,o} \right) \leftarrow Q\left( {s,a,o} \right) + \alpha \left[ {r\left( {s,a,o} \right) + \gamma V\left( {s'} \right) - Q\left( {s,a,o} \right)} \right]
\end{equation}
where $ \alpha \in \left(0,1 \right]  $ is the learning rate, $ V\left(s\right)$ is the state function, and $ s' $ is the state of the next time step. For deterministic action selection policies \cite{littman1994markov}, $ V\left(s\right) $ is calculated as follows:
\begin{equation}\label{euq2}
V\left( s \right) = \mathop {\max }\limits_{a \in \mathcal{A}} \mathop {\min }\limits_{o \in \mathcal{O}} Q\left( {s,a,o} \right)
\end{equation}
Action selection strategies such as $ \epsilon-greedy $ and Boltzmann distribution \cite{sutton2018reinforcement} can also be used in Minimax-Q, and the optimal policy is:
\begin{equation}\label{euq3}
{\pi ^*}\left( s \right) = \arg \mathop {\max }\limits_{a \in \mathcal{A}} \mathop {\min }\limits_{o \in \mathcal{O}} {Q^*}\left( {s,a,o} \right)
\end{equation}

Besides, Minimax-SARSA is an on-policy variant of Minimax-Q, whose performance strongly depends on the actual learning policy. The updating rule is similar to \eqref{euq1} except that $ V\left(s' \right)  $ is replaced by the actual joint action value:
\begin{equation}\label{equ4}
V\left( s' \right) =  Q\left( {s',a',o'} \right)
\end{equation}
where $ a' \in \mathcal{A} $ and $ o'  \in \mathcal{O} $ represent the actions of the next time step of the agent and the opponent, respectively.

One drawback of Minimax-Q and Minimax-SARSA is that only one state value is updated per iteration, which is inefficient. To address this problem, Banerjee et al. \cite{Banerjee2001Fast} have proposed Minimax-Q($ \lambda $) and Minimax-SARSA($ \lambda $), which combines RL algorithms and the eligibility trace technique. Minimax-Q($\lambda$) and Minimax-SARSA($\lambda$) are considered to accelerate learning through temporal generalization because the previously accessed $ (s, a, o) $ tuples in the trajectory are also updated in each iteration.

Furthermore, trying to speed up the learning process, a number of Minimax-Q variants with exploration heuristics have been presented. Ribeiro et al. \cite{Ribeiro2002Experience} have proposed Minimax-QS, in which a single experience can be used to update a couple of pairs through spatial generalization. The spreading function ${\sigma _t}\left( {s,a,o,{s_i},{a_i},{o_i}} \right) \in \left[ {0,1} \right]$ defines the similarity among state-action pairs. In \cite{Ribeiro2002Experience}, $ {\sigma _t}\left( {s,a,o,{s_i},{a_i},{o_i}} \right) = {g_t}\left( {s,{s_i}} \right)\delta \left( {a,{\rm{ }}{a_i}} \right)\delta \left( {o,{o_i}} \right) $, where $ g_t $ is the state similarity function, and $\delta$ is the Kronecker delta function. For each update, the Q-value of the tuple $\left(  {s_i},{a_i},{o_i}\right)  $ is updated simultaneously according to the similarity degree to the tuple $\left( s, a, o \right) $. Bianchi et al. \cite{Bianchi2007Heuristic, Bianchi2014Heuristically} have proposed HAMMQ, which uses a heuristic function to guide action selection. However, the handcrafted heuristics in HAMMQ are state specific without generality. In other words, states not covered by the heuristics cannot be updated under guidance.

The above Minimax-Q like algorithms share an disadvantage that they use the concept of Nash equilibrium by assuming that the opponent always follows the optimal action selection policy, while these algorithms follow a conservative policy. However, the opponent may follow a suboptimal policy (not always choosing the best action) or other kinds of policies. Then, the learned response policies may not be optimal.

The above problem can be partially addressed by tracing other players' policies during online interactions. In this situation, the goal of Markov games is to obtain the optimal payoff in the presence of other agents. Claus et al. \cite{claus1998dynamics} were the first to introduce other agents' models into fully cooperative tasks. Uther et al. \cite{uther1997adversarial} have incorporated fictitious play \cite{shoham09a} in games into policy learning in fully competitive tasks. Weinberg et al. \cite{weinberg2004best} have introduced the NSCP learner to general-sum Markov games to get the best response policy by inferring the model of the non-stationary opponents. Hernandez-Leal et al. \cite{hernandezleal2014a} have proposed a model-based algorithm named MDP-CL for repeated games, which leans the opponent's dynamics in the form of MDP and detects the opponent's strategy every fixed number of episodes. In all the studies reviewed here, other players' models (or opponent models) are constructed as probabilistic models. Specifically, they record the number of times that the actions have been executed by other players in given states. However, for states that have never been accessed but have features similar to the recorded ones, these approaches can not predict the behaviors of other players. Moreover, the state-action memory grows exponentially with the number of states and actions, which is infeasible when handling complex problems. In this paper, we construct an opponent model with an neural network. Combing with the representationc technique and \textit{matching mechanism} \cite{wilson1998generalization} of XCS, our approach can predict the opponent's behavior in the states with similar features.

\subsection{NNbRL for Markov games}

One efficient way of addressing the huge and/or continuous state space is to introduce neural networks as the function approximators. Mnih et al. \cite{mnih2016asynchronous} have presented an asynchronous framework for deep RL and proposed the asynchronous advantage actor-critic (A3C) for continuous control problems. In order to learn efficiently in rich domains, Heess et al. \cite{heess2017emergence} have proposed the distributed proximal policy optimization (DPPO) algorithm, in which data collections are distributed over workers. Similar ideas can be found in proximal policy optimization (PPO) \cite{schulman2017proximal}. However, these NNbRL algorithms require the participation of multiple parallel learners and therefore requires more computing resources.

A variety of previous work have investigated the incorporation of  other players' models into NNbRL algorithms. Lowe et al. \cite{lowe2017multi} have extended DDPG \cite{lillicrap2016continuous} to MAS and proposed MADDPG for multi-agent cooperation and competition. He et al. \cite{he2016opponent} have proposed to incorporate the mix-of-experts architecture to deep Q-network (DQN) \cite{mnih2015human} for different types of opponents and presented the DRON. One limitation of DRON is that it can not response optimally against a particular type of opponent. Similar ideas can also be found in DPIQN \cite{DPIQN}, in which the agent learns the policy features of the opponent as auxiliary tasks and incorporates the learned features into the Q-network. One difference between DPIQN and our work is that we learn a separately explicit opponent model not only for action selection but also for the evolution of action selection rules. Raileanu et al. \cite{raileanu2018modeling} have proposed self other-modeling (SOM) to infer the other players' hidden goals from their behaviors and use these estimates to choose actions. However, SOM requires longer training time to infer the opponents' goals.

For opponents switching between several stationary strategies, it is an effective way to store the learned policies as heuristics. Then in the application phase, the agent detects whether the opponent's policy changes and reuses the optimal response policy. Hernandez-Leal et al. \cite{hernandezleal2016identifying} have combined Bayesian policy reuse (BPR) \cite{rosman2016bayesian} and R-max in repeated games and proposed a model-based algorithm called BPR+, in which different opponent's policies are considered as different tasks in BPR. Besides, Bayes-Pepper \cite{hernandezleal2017towards} has been proposed for Markov games, which introduces the intra-belief to help the agent adapt its policy when the opponent's policy changes within an episode. As a table-based algorithm, Bayes-Pepper may be infeasible when handling complex scenarios. As a solution, Hao et al. \cite{NIPS2018_7374} have proposed Deep BRP+ that integrates BPR+ \cite{hernandezleal2016identifying} with DQN \cite{mnih2015human}, in which the policy disillusion technique is used as the policy library in BPR+. One deficiency of Deep BPR+ is that it can only trace the opponents who use random switching policies. However, in practice, the opponent may choose a policy based on the interaction history of the agents. As a solution, Bayes-ToMoP \cite{yang2019towards} has been proposed assuming that the opponents may also have the reasoning ability, e.g, use BPR. However, Bayes-ToMoP is inefficient when handling tasks with different optimal return. In this paper, we only deals with the opponent that evolves its policy simultaneously with the agent rather than switches between known policies.



\subsection{EXtended classifier system (XCS)} \label{the xcs}


\begin{figure}
	\centering
	\includegraphics[width=0.6\textwidth]{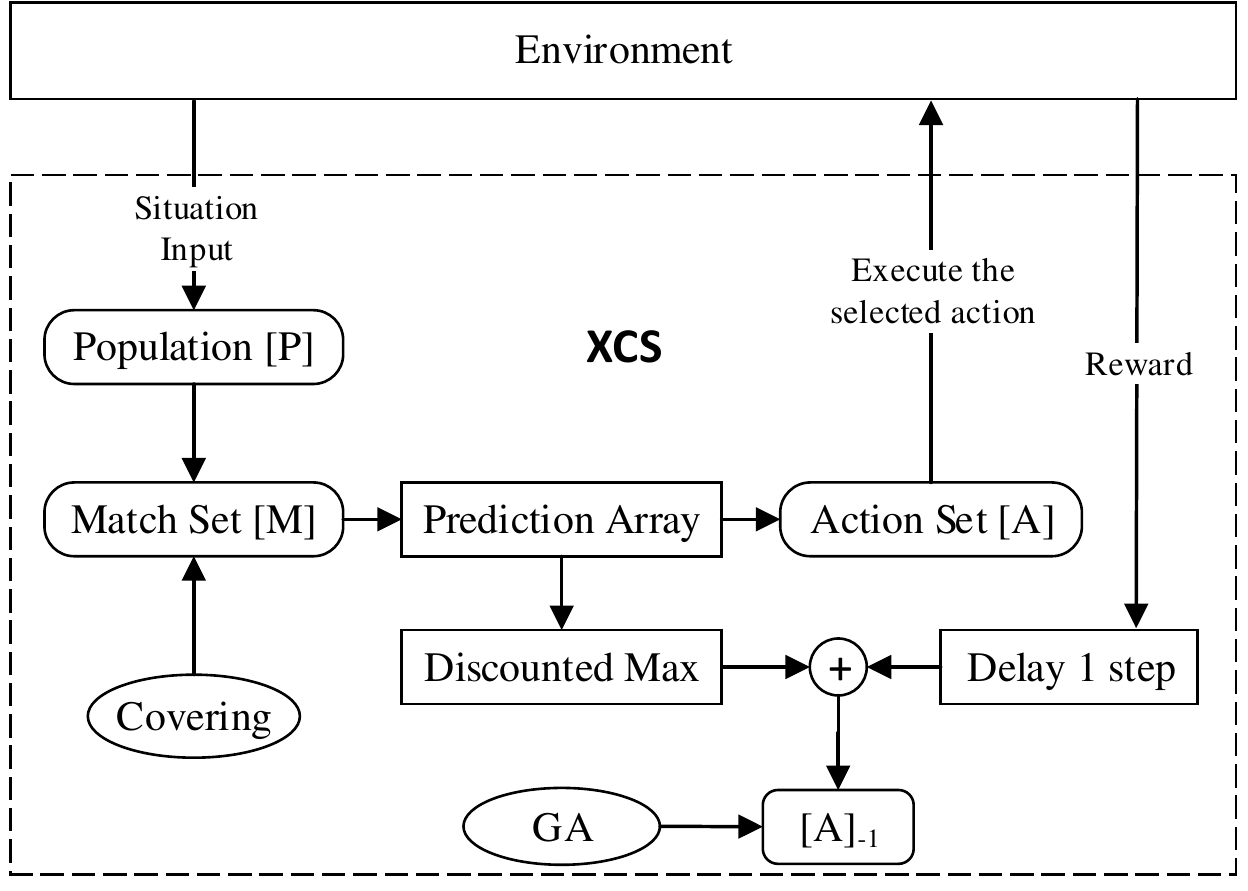}
	\caption{An overview of XCS.}
	\label{xcs}   
\end{figure}

XCS \cite{wilson1995classifier, wilson1998generalization} is an accuracy based Michigan-style LCS \cite{Holland1992Adaptation}, in which the strength of the classifier is replaced by the payoff prediction $ p $, the prediction error $\epsilon$, and the fitness $ F $. Besides, each classifier $ cl $ also keeps additional parameters to support learning in XCS, e.g., the action set size $ as $ estimates the averaged size of the action set that contains $ cl $, the time stamp $ ts $ indicates the time-step of the last GA in the action set that contains $ cl $, the experience $ exp $ specifies the number of times $ cl $ has been selected in an action set, etc. 

We note that such a classifier $ cl $ is also called a rule in the XCS literature. Besides, the state $ s \in \mathcal{S} $ in RL is called the situation in XCS (after coding)\footnote{In this paper, we use the term classifier and rule alternatively as needed. We use $ s $ to refer to a given state or its corresponding situation indiscriminately.}.

The overview of XCS is shown in Fig. \ref{xcs}. When XCS receives an input situation $ s $ from the environment, a match set [M] is formed, containing every classifier in [P] whose condition matches $ s $. The \textit{covering} mechanism is necessary if the number of actions in [M] is less than a predefined threshold $\theta_{mna}$, which guarantees the candidate actions for selection are sufficient. Then, new classifiers with conditions matched with $s$ will be created for those missing actions with predefined predicted payoff $p$ along with $\epsilon$ and $F$. The \textit{system prediction} of an action is a fitness-weighted average of the predicted payoff of all classifiers in [M] advocating that action. The prediction array in Fig. \ref{xcs} contains the \textit{system predictions} of all candidate actions. Moreover, the action with the highest \textit{system prediction} is selected to execute and the classifiers in [M] advocating that action forms the action set [A]. After the selected action is performed and the actual reward is obtained, the prediction error is calculated and used to update all the relevant rules in the previous time step's action set $ [\rm{A}]_{-1} $.

GA is responsible for classifier generalization in XCS. Specifically, GA is triggered if the average time step of classifiers in the action set exceeds the threshold $ \theta_{GA} $.
Two parent classifiers are selected from $ [\rm{A}] $, with probability proportionate to their fitness $ F $. Then, the offspring classifiers which can still match the current input are created through the crossover and mutation operators. Finally, parameters of the offspring classifiers are initiated. The numerosity $ num $ and experience $ exp $ are initialized to 1, and the payoff prediction $ p $, the prediction error $\epsilon$, and the fitness $ F $ are initialized in a certain proportion of the parents'.


After creating new classifiers in GA or updating a classifier in the action set, the \textit{subsumption mechanism} is triggered. \textit{GA subsumption} checks the offspring classifiers to see whether their conditions can be subsumed by the classifier that is accurate enough ($ cl.\epsilon < \epsilon_{0} $) and sufficiently experienced ($ cl.exp > \theta_{sub} $)\footnote{We use a dot to refer to a parameter in a classifier.}.  \textit{Action set subsumption} finds the most general classifier that is accurate and sufficiently experienced in [A] and tries to subsume the less accurate one. After applying the \textit{subsumption mechanism}, the numerosity parameter $ num $ of the subsumer (named \textit{macroclassifier}) is increased. In other words, one \textit{macroclassifier} represents multiple regular classifiers with the same parameters. \textit{Macroclasifier} and \textit{subsumption} are two major mechanisms to speed up matching classifiers with an input and evolving maximally general classifiers.

In order to maintain a fixed-size population, the classifiers with low fitness are deleted.
Specifically, sufficiently experienced ($ cl.exp>\theta_{del} $) classifiers with low average fitness ($ cl.F/cl.num<\delta \cdot F_{average} $), large action set size, and large numerosity are more likely to be deleted \cite{wilson1995classifier}.

\section{The HAMXCS algorithm}

\begin{figure}
	\centering
	\includegraphics[width=1.0\textwidth]{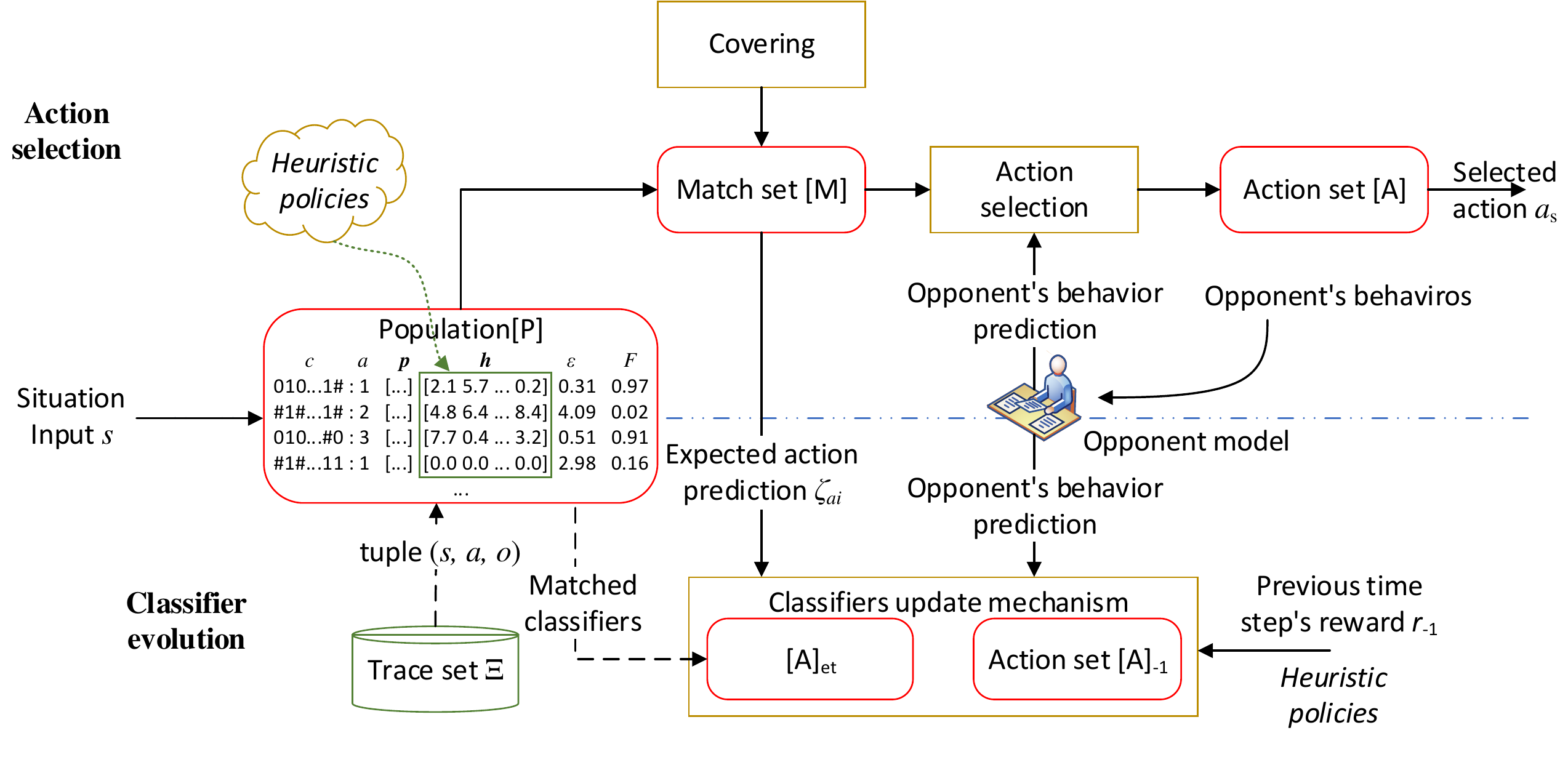}
	\caption{An overview of HAMXCS. The upper part describes the process of action selection and the lower part indicates the process of evolving classifiers. Data modules are represented by the rounded rectangles and the main function models are described in rectangles.}
	\label{pre}   
\end{figure}

The overview of HAMXCS is shown in Fig. \ref{pre}. The opponent model goes through the whole process of action selection and rule evolution. Moreover, the heuristic variable is incorporated into the classifier representation and updated in the learning part. Besides, we design an accuracy-based eligibility trace mechanism for XCS framework to speed up policy learning, which is shown by the dotted lines in the lower part of Fig. \ref{pre}. 

\label{HAMXCS}

\subsection{Classifiers representation in HAMXCS}
\label{representation hamxcs}
The classifiers are the basic units in HAMXCS which make up the population [P] and are used for policy learning. We extend the classifier representation in XCS to Markov games scenarios by integrating heuristics and opponent information. Specifically, the classifier $ cl $ in HAMXCS maps the condition $ c $ and the agent's action $ a $ to a tuple $ <\bm{p}, \bm{h}_j, \epsilon, F> $ as follows:
\begin{equation}
<c,a> \rightarrow <\bm{p},\bm{h}_j, \epsilon, F>
\end{equation}
where $ \bm{p} $ is the payoff prediction vector, $ \bm{h}_j\left( j=1,2,\dots,n \right) $ is the informed component which represents the $ j $-th heuristic vector, $\epsilon$ is the prediction error, and $ F $ is the fitness of the classifier. The element $ p[o] $ in $ \bm{p} $ indicates the predicted payoff if the HAMXCS agent takes action $ a $ while the opponent executes action $ o $ in the given situation. The heuristic value $ h_j[o] $ in $ \bm{h}_j \left( j=1,2,\dots,n\right) $ describes the degree to which the $ j$-th heuristic advises the agent to choose action $ a $ when the opponent selects action $ o $. For example, assuming that there are 4 possible actions for the agent and the first heuristic vector $ \bm{h}_1$ in a classifier is [1.0, 2.5, 10.5, 0.3]. In this case, $ \bm{h}_1$ gives more priority to the 3rd action, which will influence the action selection in Section \ref{action selection}. Besides, $ \bm{h}_j\left( j=1,2,\dots,n \right) $ will be updated according to the corresponding heuristic policy, which is detailed in Section \ref{Update the heuristic component}. In the following descriptions, we denote $ \bm{h} $ as the heuristic vector and $ h[o] $ as the heuristic value in $ \bm{h} $ unless it's necessary to clarify that there are multiple heuristics $ \left(j>1 \right)  $ in the $ cl $.

The condition $ c $ in $ cl $ is represented in the ternary alphabet $ \left\lbrace 0, 1, \#\right\rbrace $, in which the wildcard $ \# $ can be either 0 or 1. In contrast, the situation $ s $ (i.e., the state in RL) is binary coded, thus each bit is 0 or 1.

The \textit{matching mechanism} guarantees a given situation $ s $ can be matched with a variety of classifiers with different levels of generalized conditions. Specifically, when the HAMXCS agent receives a situation $ s $ from the environment, it matches the condition $ c $ of each classifier in [P] by checking the no-wildcard components. After that, the selected classifiers in [P] forms the match set [M]. For example, suppose 1\#010 and 111\#\# are the conditions of two classifiers in [P]. Then, only the first classifier matches with an input 11010. We note that the match set [M] consists of all the action sets that suggest the selection of different candidate actions.

\subsection{Opponent model construction and update}
\label{opponent model}
In HAMXCS, the opponent model is constructed from scratch and updated using the observed opponent's behaviors during online interactions. In the application phase, the learned opponent model is responsible for predicting the opponent's action in given states. In the literature \cite{XOMQ, claus1998dynamics, uther1997adversarial, hernandezleal2014a}, opponent models are constructed by simply counting the opponent's action execution frequencies. The main drawback of these approaches is that it can only predict the opponent's action in visited states. However, in practical application, some states are similar in feature representation. As a result, the opponents might behave similarly in these states.

To address this problem, we use a neural network to approximate the opponent's actual policy. Specifically, we first collect the opponent's behaviors episodically and then use the collected data to update the opponent model. Suppose that the opponent's behavior sequence is $ \left(s_0, o_0, s_1, o_1, \dots, s_t, o_t \right)  $ during the time period $ t $, the opponent model $\tau$ is updated by maximizing the log probability of generating the sampling sequence. However, different behavior sequences may vary greatly because the opponent is learning concurrently. Therefore, the entropy of the predicted policy $ E\left(\tau \right)  $ is added to the loss function to avoid over-fitting. As a result, the loss function is presented as follows:
\begin{equation}
Loss\left(\phi \right)=-\mathbb{E}_{s_i,o_i}\left[ \log  \tau\left(o_i|s_i\right)+\eta E\left(\tau \right)  \right]  
\end{equation}
where $\phi$ represents the parameters of the opponent model, $ \tau\left(o_i|s_i\right) $ is the probability of the opponent taking action $ o_i $ in state $ s_i $ predicted by the opponent model, and $\eta \in [0,1]$ is the entropy parameter.

One advantage of our opponent model is that benefiting from the $ \left\lbrace 0, 1, \#\right\rbrace $ condition representation and \textit{matching mechanism} in Section \ref{representation hamxcs}, the opponent model can predict the opponent's action in the given states which are similar to the previously accessed. Moreover, the prediction is used for action selection in Section \ref{action selection} and classifiers evolution in Section \ref{update calssifiers} and \ref{update classifiers in aet}. Another advantage of the opponent model is that it might allow the agent to learn the opponent's hidden behavior features.


\subsection{Action selection with heuristics and the opponent model}
\label{action selection}
\subsubsection{Single heuristic for policy learning}
\label{single h}
Action selection strategies for SAS are also applicable to HAMXCS, such as the $\epsilon-greedy$ and the adapted Boltzmann distribution \cite{van2011insights}. After the match set [M] is formed, HAMXCS tries to select the optimal response action combing the opponent model and heuristics in the given situation $ s $. Specifically, we first estimate the fitness-weighted average prediction $ \bm{P}_{s,a_i} $ for each candidate action $ a_i $ in [M]:
\begin{equation}\label{action predcition}
\bm{P}_{s,a_i} = \frac{{\sum\nolimits_{cl.a = {a_i} \wedge cl \in [\rm{M}]} {cl.\bm{p} \cdot cl.F} }}{{\sum\nolimits_{cl.a = {a_i} \wedge cl \in [\rm{M}]} {cl.F} }}
\end{equation}
where $cl.\bm{p}$ and $cl.F$ denote the payoff prediction vector and the fitness value of the classifier $cl$, respectively. The value of $ P_{s,a_i} [o]$ in $ \bm{P}_{s,a_i} $ indicates the payoff prediction of HAMXCS after the agent and the opponent execute action $ a_i $ and $ o $, respectively. 

Then, the fitness-weighted average heuristic (FAH) $ \bm{H}_{s,a_i} $ of each candidate action is calculated similarly:
\begin{equation}\label{hfunction}
\bm{H}_{s,a_i} = \frac{{\sum\nolimits_{cl.a = {a_i} \wedge cl \in [\rm{M}]} {cl.\bm{h} \cdot cl.F} }}{{\sum\nolimits_{cl.a = {a_i} \wedge cl \in [\rm{M}]} {cl.F} }}
\end{equation}
where $cl.\bm{h}$ indicates the heuristic vector of $ cl $. The element $ H_{s,a_i} [o]$ in $ \bm{H}_{s,a_i} $ describes the degree to which the heuristic suggests HAMXCS selects action $ a_i $ based on the accuracy of the advocated classifiers and the opponent's corresponding action $ o $. 

Next, combined with the current opponent model $ \tau $ and $ \bm{H}_{s,a_i} $, the expected prediction with heuristics (EPH) $ \xi_{a_i}\left( s\right)  $ for each action $ a_i \in \mathcal{A}$ is presented as follows: 
\begin{equation}\label{action with h}
\xi_{a_{i}} \left( s\right) = \kappa \cdot H_{s,a_i} \left[ o_s\right] + \sum \limits_{{o\in \mathcal{O}}}	\tau \left(o|s\right)  \cdot P_{s,a_i} \left[ o\right] 
\end{equation}
where $ o_s $ is the executed action by the opponent and $\kappa \in \mathbb{R}$ is a parameter to weight FAH. EPH describes the expected payoff in the current situation $ s $ after taking account of the opponent model $\tau$ and the heuristic. Finally, action selection strategies can be applied to HAMXCS. Take the $ \epsilon-greedy $ exploration policy as an example, the action selection rule can be presented as follows:
\begin{equation}
\pi \left( s \right) = \left\{ \begin{array}{l}
\arg \mathop {\max }\limits_{a \in \mathcal{A}}  \xi_{a}\left( s\right)
\quad{\rm{if}} \ prob \ge \varepsilon \\
{a_{random}}\qquad {\rm{otherwise}}
\end{array} \right.
\end{equation}
where $ prob $ is a random variable drawn from a uniform distribution over $ [0,1] $. 

In contrast to the QbRL algorithms such as Minimax-Q, our approach of action selection takes into account the opponent's behaviors. Specifically, if the opponent never executes action $ o $, the corresponding payoff prediction is ignored by the opponent model. Another advantage of our method is that the heuristic function in (\ref{hfunction}) considers the accuracy of each classifier. In detail, the higher the accuracy of the classifier, the stronger the guidance of the heuristic.

In practice, the prior knowledge is often only a rough heuristic policy and not covers the whole state space \cite{zhang2020kogun}. Leveraging the \textit{matching mechanism} and generalization capability of HAMXCS, situations that match with the informed classifiers (i.e., classifiers with non-zero heuristic $ cl.\bm{h} $) can also be guided during action selection. For example, the heuristic policy indicates that if the agent is located in situation 1001, it is recommended to select action $ a_0 $. Assume that HAMXCS has a classifier $ cl_0 $ with condition 1\#\#1 and action $ a_0 $, and then it is selected into the action set [A] and $cl_0. \bm{h} $ is updated as detailed in Section \ref{Update the heuristic component}. After that, the HAMXCS agent can also benefit from $cl_0. \bm{h} $ when it is in situations match 1\#\#1 while ignored by the heuristic policy, such as 1011.


\subsubsection{Multiple heuristics for policy learning}
Another situation is that there are more than one heuristics in $ cl $. In this case, action selection strategies in Section \ref{single h} are also applicable. An intuitive idea is to use the weighted sum of the FAHs as the final heuristic value in (\ref{action with h}). The weights represent the importance of the corresponding heuristics. The limitation of this approach is that it only describes the linear relationships of the heuristics and it is not easy to determine the weights.

Another feasible way is to introduce the Pareto optimality \cite{shoham09a} to action selection. Assume that there are $ n $ heuristic policies identified by $ j\in \mathcal{T} \equiv \left\lbrace1,2,\dots,n \right\rbrace $. Correspondingly, EPHs related to action $ a \in \mathcal{A}$ is denoted as $ \xi_{a}^j \left( j=1,2,\dots,n\right) $. We take the idea of Pareto optimality and formalize the following definitions. 

\begin{definition}\label{def1}
	(Pareto action domination) For $ a_i,a_k \in \mathcal{A} $, action $ a_i $ Pareto dominates action $ a_k $ in given state $ s \in \mathcal{S}$, if for all $ j \in \mathcal{T} $, $ \xi_{a_i}^j \left(s \right) \geq \xi_{a_k}^j\left( s\right) $, and there exists some $ m \in \mathcal{T}$ for which $ \xi_{a_i}^m \left( s\right) > \xi_{a_k}^m\left( s\right) $.
\end{definition}

In other words, a Pareto dominated action can only have a larger EPH value than other candidates for all heuristic policies. In this case, Pareto action domination gives us a partial ordering over the agent's actions. Then, we get a set of incomparable optima solutions rather than a single best response action with Definition \ref{def2}.

\begin{definition}\label{def2}
	(Pareto optimal action) Action $a_{p} $ is a Pareto optimal action, if there does not exist another action $ a\in \mathcal{A} $ that Pareto action dominates $ a_p $.
\end{definition}

For classifiers with many heuristics, the Pareto front approximated approaches \cite{LiuMultiobjective} can be applied to solve the Pareto optimal actions. Finally, we obtain a Pareto optimal action set and then randomly select an action from the set for execution. In contrast to the weighted sum method, it is suitable for scenarios where it is difficult to identify the relative weight of the heuristics. Besides, it can also represent the non-linear relationships between the heuristics.

\subsection{Classifiers update in $ [\rm{A}]_{-1} $}
\label{update calssifiers}
Once the action $ a_s $ is selected in Section \ref{action selection}, the classifiers that advocates $ a_s $ in [M] forms the action set [A]. After $ a_s $ is executed, HAMXCS receives the reward $ r $ from the environment, which is used to update the parameters of the classifiers in the last time step's action set $ [\rm{A}]_{-1} $. 



\subsubsection{Update the payoff components}
The expected action prediction (EAP) $ \zeta_{a} \left(s \right) $ is the payoff prediction when HAMXCS selects action $ a $ while the opponent follows the estimated policy $ \tau $. Formally, we have:
\begin{equation}\label{EAP}
\zeta_{a} \left(s \right) = \sum \limits_{{o \in \mathcal{O}}}	\tau \left(o|s\right)  \cdot P_{s,a} \left[ o\right]
\end{equation}
where $ P_{s,a}\left[ o\right] $ is the payoff prediction value in (\ref{action predcition}).

The target prediction $ P_t\left( s\right)   $ (abbreviated as $ P_t $ for simplicity) indicates the expected total discounted future payoff of $ [\rm{A}]_{-1} $, resulting from taking the selected action $ a_{-1} $ in the previous situation $ s_{-1} $ and continuing the optimal policy thereafter. In other words, $ P_t $ is the sum of reward $ r_{-1} $ and the discounted maximal EAP:
\begin{equation} \label{target prediction}
P_t = r_{-1} + \gamma \max \limits_{a \in \mathcal{A}} \zeta_{a}\left(s \right)
\end{equation}
where $ \gamma \in \left[ 0,1\right)$ is the discount factor. Note that the obtained reward is also related to the opponent' action $ o_{-1} $ in the previous time step. As the result, only the payoff prediction corresponding to action $ o_{-1}  $ is updated:
\begin{equation}\label{}
cl.p[o_{-1} ] \leftarrow cl.p[o_{-1} ]+\beta_{1}\left(P_t-cl.p[o_{-1} ] \right) 
\end{equation}
where $\beta_1 \in  \left(0,1 \right] $ is a learning rate. Please note that the heuristics only affect action selection in (\ref{action with h}) without participating in the payoff prediction update. Therefore, the introduction of heuristics will only affect the learning efficiency but not the convergence of the algorithm itself.

The prediction error $ \epsilon $ describes the difference between $ P_t $ and the expected payoff after executing action $ a_{-1} $. Since the opponent model $ \tau $ and $ \bm{p} $ are constantly updated, $\epsilon$ is updated accordingly. We use the expected classifier prediction (ECP) $ cl.\zeta $ to denote the predicted payoff that the classifier $ cl $ receives under the current opponent model:
\begin{equation}
cl.\zeta = \sum \limits_{{o \in \mathcal{O}}} \tau \left(o|s_{-1} \right) \cdot cl.p\left[ o\right]
\end{equation}
Where $ \tau \left(o|s_{-1} \right) $ is the prediction of the opponent model in $ s_{-1} $. Then, the prediction error of each classifier is updated as follows:
\begin{equation}
cl.\epsilon\leftarrow cl.\epsilon + \beta_2\left( \left|P_t-cl.\zeta\right| - cl.\epsilon\right) 
\end{equation}
where $\beta_2 \in  \left(0,1 \right] $ is a learning rate.



The fitness $ F $ reflects the accuracy of a classifier, and it is updated in the same way with vanilla XCS:
\begin{equation}\label{update fitness}
cl.F \leftarrow cl.F+\beta_3\left(\frac{{cl.k \cdot cl.num}}{{\sum\nolimits_{x \in [\rm{A}]_{-1}} {x.k \cdot x.num} }}-cl.F\right)
\end{equation}
where $ x $ is the classifier in $ [\rm{A}]_{-1} $, $\beta_3 \in  \left(0,1 \right]$ is a learning rate, and $ cl.k $ denotes the current absolute accuracy of the classifier:
\begin{equation}\label{}
cl.k = \left\{ \begin{array}{l}
1
\qquad\qquad\,\;\;\ {\rm{if}} \ cl.\epsilon  <\epsilon_{0} \\
{\alpha\left(\frac{cl.\epsilon}{\epsilon_{0}}\right)^{-\nu}  }\quad {\rm{otherwise}}
\end{array} \right.
\end{equation}
where $\alpha$ and $\nu$ are real values that further differentiate the accuracy of classifiers. $ \epsilon_{0} $ is the error threshold that indicates the maximal error tolerance.

\subsubsection{Update the heuristic component}
\label{Update the heuristic component}
The update of the heuristic vector $ \bm{h} $ is also related to the opponent's last action $ o_{-1} $, because the FAH corresponding to $ o_{-1} $ suggests the selected action $ a_{-1} $ in (\ref{action with h}). Note that according to the action selection policy in Section \ref{action selection}, the heuristic value should be positive to provide guidance for learning. As a result, for situation $ s_{-1} $ in the last time step, the heuristic value $ h[o_{-1}] $ is updated as follows:
\begin{equation}
h[o_{-1}]\leftarrow \max_{a \in \mathcal{A}}P_{s_{-1},a} [o_{-1}]- P_{s_{-1},a_h} [o_{-1}] + \rho
\end{equation}
where $ P_{s_{-1},a} [o_{-1}] $ is the fitness-weighted average prediction value in (\ref{action predcition}) in $ s_{-1} $, $ a_h $ is the action suggested by the heuristic policy, and $ \rho \in \mathbb{R}$ is	 a variable that affects the update magnitude. We use $ \bm{P}_{s_{-1},a} $ to update $ \bm{h} $ because it absolutely indicates the payoff. Besides, in this way, the heuristic guides action selection without causing large errors in the true payoff prediction, EAP.

In the following theorem, we show the upper bound of the EAP error caused by introducing FAH when selecting action greedily. Given a situation $ s \in \mathcal{S} $ and the opponent model $ \tau $, the optimal policy $ \pi^* $ is defined as:
\begin{equation}\label{optimal action}
\pi^*\left( s\right) =\arg\max \limits_{a \in \mathcal{A}}\zeta_{a} \left(s \right) 
\end{equation}
The greedy policy $\tilde{\pi}  $ with FAH can be presented as follows:

\begin{equation}\label{EPH action}
\tilde{\pi} \left( s\right)  =\arg\max \limits_{a \in \mathcal{A}}\xi_{a}\left(s \right) 
\end{equation}
Formally, we define the EAP error $ Er \left( s\right) $ for situation $ s \in \mathcal{S} $ as follows:
\begin{equation}
Er \left( s\right) = \zeta_{ \pi^*} \left(s \right) -\zeta_{\tilde{\pi}} \left(s \right)
\end{equation}
where $ \zeta_{ \pi^*} \left(s \right) $ and $\zeta_{\tilde{\pi}} \left(s \right) $ are the EAPs of actions got from $ \pi^* $ and $\tilde{\pi}  $, respectively.

\begin{theorem}\label{theo1}
	For HAMXCS in zero-sum Markov games with finite states and actions, bounded rewards, the discounted factor $ \gamma \in \left[ 0,1\right)  $, and the opponent model $\tau$, the upper bound of the EAP error $ Er\left( s\right)  $ resulted by introducing FAH $ H_{s,a}[\cdot] \in \left[\mathcal{V}_{min},\mathcal{V}_{max} \right]  $ for each situation $ s \in \mathcal{S} $ is:
	\begin{equation}
	Er\left( s\right) \leq \kappa \left(\mathcal{V}_{max}-\mathcal{V}_{min} \right) 
	\end{equation}
\end{theorem}

\newproof{pot}{Proof of Theorem \ref{theo1}}
\begin{pot}
	Assume that there exist a situation $ s_m $ that achieves the maximum EAP error: $ \exists s_m \in \mathcal{S}, \forall s \in \mathcal{S}, Er(s_m) \geq Er(s) $. For situation $ s_m $ consider an optimal action, $ a^*=\pi^*(s_m) $, and the action of $ \tilde{\pi} $, $  \tilde{a}= \tilde{\pi}(s_m) $. Because $ \tilde{\pi} $ is a greedy policy of EPH, according to (\ref{action with h}), we have:
	\begin{equation}\label{proof}
	\begin{split}
	&\kappa \cdot H_{ s_m,\tilde{a}} \left[ o\right] + \xi_{ \tilde{a}}\left( s_m\right)  \geq \kappa \cdot H_{ s_m, a^*} \left[ o\right] + \xi_{  a^*}\left(s_m \right) \\
	&Er \left( s_m\right) = \zeta_{ a^*} \left(s_m \right) -\zeta_{ \tilde{a}} \left(s_m \right) \leq \kappa \left(H_{s_m, \tilde{a}} \left[ o\right]- H_{s_m,  a^*} \left[ o\right]\right) 
	\end{split}
	\end{equation}
	where $ o $ is the selected action of the opponent. From (\ref{optimal action}), we have 
	$ \zeta_{ a^*}\left( s_m\right) \geq \zeta_{  \tilde{a}}\left( s_m\right)$. Combing (\ref{proof}), we have $ H_{ s_m, a^*} \left[ o\right] \leq H_{ s_m,\tilde{a}} \left[ o\right] $. Finally, with the bound of FAH $ H_{s,a}[\cdot] \in \left[\mathcal{V}_{min},\mathcal{V}_{max} \right]  $, for $ \forall s \in \mathcal{S} $, we have:
	\begin{equation}
	Er\left( s\right) \leq \kappa \left(\mathcal{V}_{max}-\mathcal{V}_{min} \right).
	\end{equation}
\end{pot}
The proof is completed.$\hfill\blacksquare$ 

After the above parameters are updated, the experience parameter $ exp $ is incremented by 1 and the action set size $ as $ is updated as the vanilla XCS. Besides, GA and the \textit{subsumption mechanism}  are executed in $ [\rm{A}]_{-1} $ as described in Section \ref{the xcs}. 

\subsection{The accuracy-based eligibility trace mechanism in HAMXCS}
\label{et}

Eligibility trace is one of the basic mechanisms of RL \cite{sutton2018reinforcement}. It records the historical states and actions during online learning. Once the agent is rewarded form the environment, the recently visited state-action values are also updated. In this paper, we introduce the accuracy-based eligibility trace into HAMXCS for further speeding up the learning process.

\subsubsection{The trace set and eligibility value}

HAMXCS maintains a trace set $\Xi$ and an eligibility trace $ e $ to record the information needed for classifiers update. The trace set $\Xi$ consists of three tuples $ \left(s,a,o \right)  $, where $ s $ is the situation, $ a $ is the agent's action, and $ o $ represents the opponent's action. The eligibility trace $ e $ records the corresponding eligibility values of the $ \left(s,a,o \right)  $ tuples in $\Xi$. As a result, in the learning component, not only the classifiers in the previous time step's action set $ [\rm{A}]_{-1} $ but the classifiers in [P] that can match the historical state-action tuples in $\Xi$ are also updated according to their fitness, which is detailed in Section \ref{update classifiers in aet}.

Once the HAMXCS agent and the opponent execute action $ a_s $ and $ o_s $ respectively in situation $ s $, tuple $  \left(s,a_s,o_s \right)$ is added to $\Xi$ if it is not included. Then, the eligibility values corresponding to situation $ s $ are updated as follows:
\begin{equation}\label{eg update}
e\left( {s,{a_i},{o_i}} \right) = \left\{ {\begin{array}{l}
	{1 \quad {a_i} = a_s,\ {o_i} = o_s}\\
	{0 \quad {\rm{otherwise}}}
	\end{array}} \right.
\end{equation}
For efficiency, $\Xi$ only maintains tuples with eligibility values significant enough. To achieve this, after each learning cycle, the eligibility value of each tuple $ \left(s,a,o \right)  $ are decayed as follows:
\begin{equation}
e\left(s,a,o \right)=\lambda \cdot \gamma \cdot \ e\left(s,a,o \right)
\end{equation}
where $\gamma \in \left[ 0,1\right)$ is the discount factor, $\lambda \in \left(0,1 \right)  $ is the trace-decay parameter, which determines the rate at which the trace falls. The tuple $ \left(s,a,o \right)  $ will be removed from $\Xi$ if its eligibility value is less than the predefined threshold ($ e\left(s,a,o \right) < \theta_{et} $). 


\subsubsection{Evolving classifiers in  $ [\rm{A}]_{et} $ based on fitness}
\label{update classifiers in aet}
\begin{figure}
	\centering
	\includegraphics[width=0.9\textwidth]{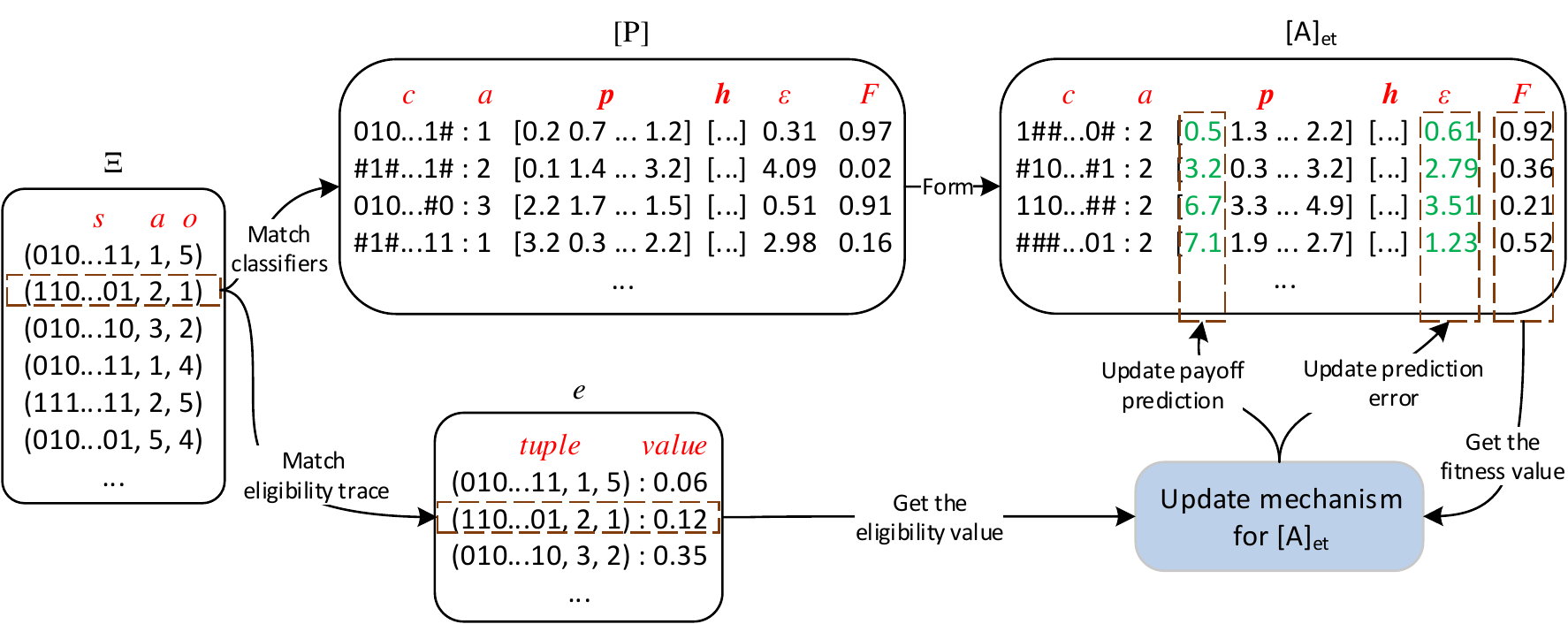}
	\caption{The workflow for classifiers update in $ [\rm{A}]_{et} $.}
	\label{aet update}       
\end{figure}

After the classifiers in $ [\rm{A}]_{-1} $ are updated, classifiers in [P] that can match the historical traces are also reinforced according to their accuracy. This can be understood as the contribution of the historical traces to the current situation. The update mechanism is shown in Fig. \ref{aet update}. 

This is achieved by matching the situation $ s $ and the action $ a $ of each record in $\Xi$ with the classifiers in [P] and forming the corresponding set $ [\rm{A}]_{et} $. Then, the classifiers in $ [\rm{A}]_{et} $ are updated depending on their fitness $ F $ and the eligibility value $ e $ of each $ \left(s,a,o \right)  $ tuple. We note that $ [\rm{A}]_{et} $ is not a real action set because the corresponding action $ a $ is not actually selected to execute. However, the classifier parameters such as the experience $ exp $, the fitness $ F $, and the action set size $ as $ should be updated only when the classifier has been selected in a real action set. Therefore, HAMXCS only updates the prediction vector $ \bm{p} $ and the prediction error $\epsilon$ of the classifiers in $ [\rm{A}]_{et} $. Besides, the updated classifiers in $ [\rm{A}]_{-1} $ are not be reinforced again. For each classifier, the update rule is as follows:
\begin{equation}\label{eg_p}
cl.p[o] \leftarrow cl.p[o]+\beta_4\left(P_t- \Phi_{-1}\right) \cdot \frac{cl.F}{\sum_{x \in [\rm{A}]_{et}}x.F}\cdot e\left( s,a,o\right)
\end{equation}
\begin{equation}\label{eg_error}
cl.\epsilon\leftarrow cl.\epsilon + \beta_5\left( \left|P_t-\Phi_{-1}\right| - cl.\epsilon\right)  \cdot \frac{cl.F}{\sum_{x \in [\rm{A}]_{et}}x.F}\cdot e\left( s,a,o\right)
\end{equation}
where $ x $ indicates the classifier in $ [\rm{A}]_{et} $, $\beta_4 \in  \left(0,1 \right]$ and $ \beta_5 \in  \left(0,1 \right] $ are learning rates, $ \left( s,a,o\right)  $ is the record in $\Xi$ and $ e\left( s,a,o\right)  $ is the corresponding eligibility value, $ P_t $ is the target prediction in (\ref{target prediction}), and $\Phi_{-1}$ is the maximum EAP in the previous time step's match set $ [\rm{M}]_{-1} $. Formally we have:
\begin{equation}
\Phi_{-1} = \max \limits_{a_{i} \in \mathcal{A}} \sum \limits_{o_{i} \in \mathcal{O}} \tau \left(o_i|s_{-1} \right) \cdot P_{s_{-1},a_i} \left[ o_{i}\right] 
\end{equation}

Note that we use the target prediction $ P_t $ and the maximum EAP $ \Phi_{-1} $ to update the classifiers in $ [\rm{A}]_{et} $. Similar ideas can be found in Minimax-Q($\lambda$) \cite{Banerjee2001Fast}, which uses error between the target Q-value and the \textit{minimax} Q-valule at $ t-1 $ to update the state-action values. The limitation of Minimax-Q($\lambda$) is that it only updates the specific state-action pairs without generality. In contrast, HAMXCS updates a set of related general classifiers for each record. Situations with similar features as $ s $ are also benefited, thus further speeding up the learning process. Besides, the classifiers in $ [\rm{A}]_{et} $ are updated based on their fitness $ F $ as shown in (\ref{eg_p}) and (\ref{eg_error}), thus ensuring the rationality and accuracy.

\section{Experiments and results}
\label{experiment}
In this section, we evaluate the performance of HAMXCS in the Hexcer soccer domain \cite{uther1997adversarial} and a modified thief-and-hunter problem \cite{yang2019towards}. We present the experimental results compared with the QbRL and the NNbRL algorithms.

\subsection{Environments settings}
\begin{figure}
	\centering
	\includegraphics[width=0.8\textwidth]{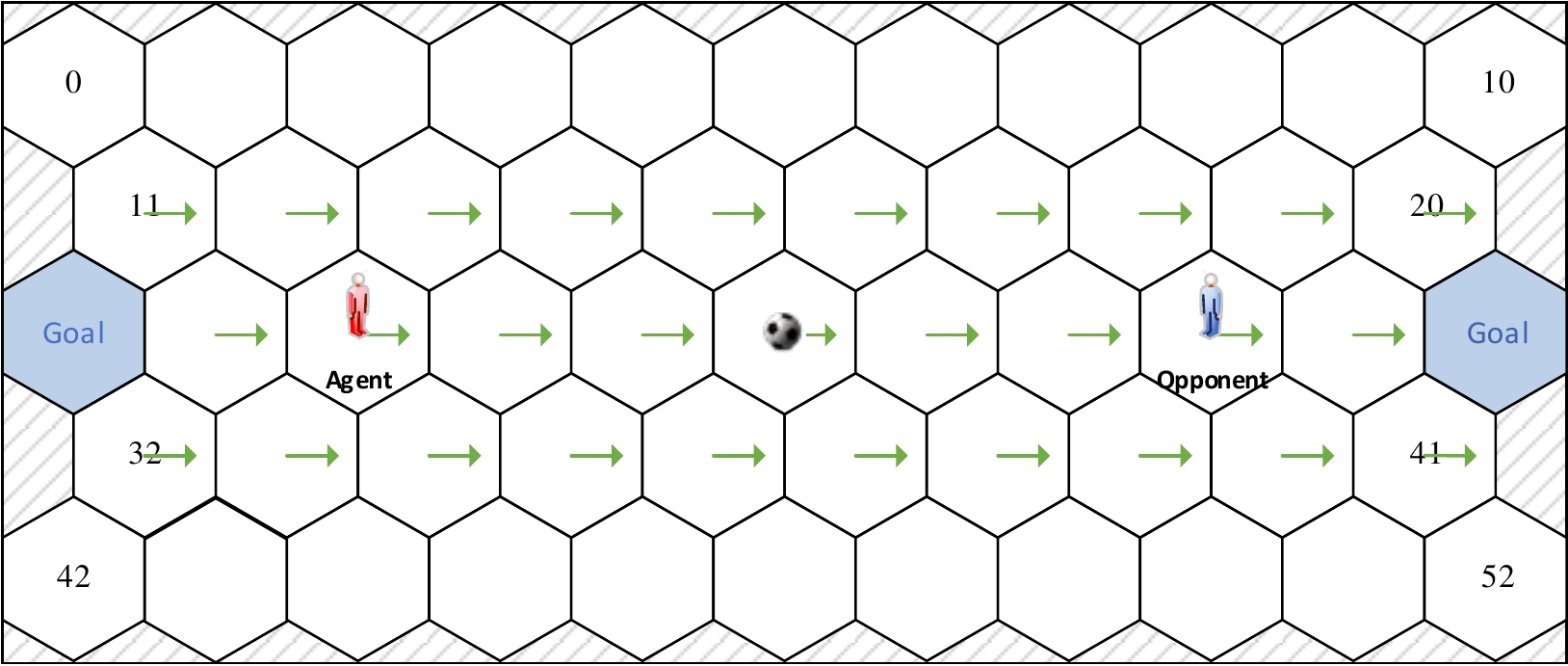}
	\caption{Hexcer environment and the heuristic policy used in the experiments. The heuristic policy is described as the right arrows.}
	\label{hexcer}       
\end{figure}

The Hexcer soccer environment is shown in Fig .\ref{hexcer}, in which the agent and the opponent are located in a grids world of connected hexagons. The initial states of the players and the ball are shown in Fig. \ref{hexcer}. Each grid can only be occupied by one player, while the ball can share a position with the player once the player gets it. The ball belongs to the player until it is taken by the opponent. When the players collide with each other, the ball will be randomly reassigned to one of the players and the corresponding move fails. Any actions that try to move the player out of Hexcer is ignored. 

The player's objective is to bring the ball into the opponent's goal area and score as many as points in a match. At each time step, a player choose an action from 7 candidate actions: $ (upper\ left, upper\ right, right, lower\ $  $right, lower\ left, left, standby) $. The game is ended when any player scores the goal or after a predefined number of time steps has passed. Once a game is ended, the position of the players and the ball will be reset.

\begin{figure}
	\centering
	\includegraphics[width=0.35\textwidth]{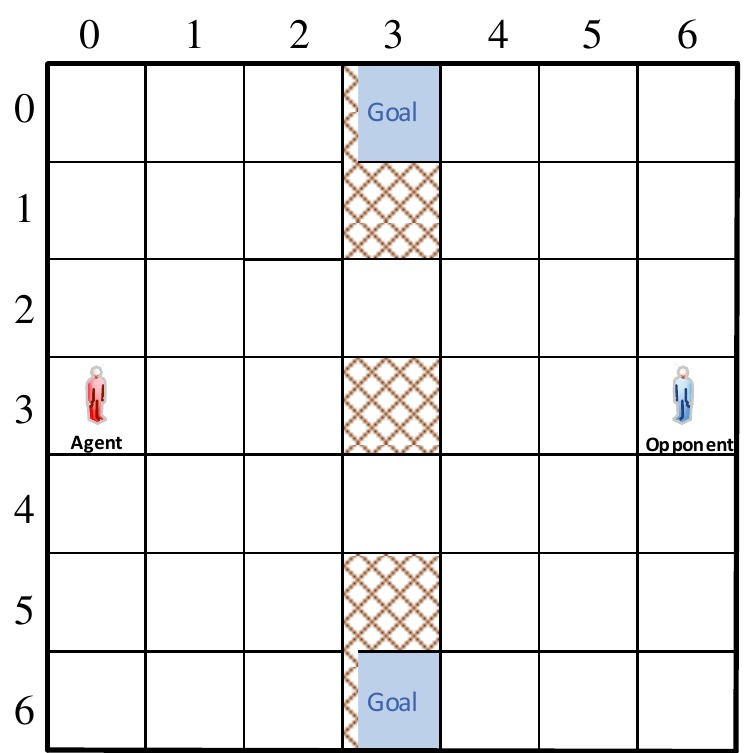}
	\caption{Thief-and-hunter environment. The shaded areas are obstacles.}
	\label{tandh}       
\end{figure}

Another testbed is the thief-and-hunter scenario. The initial states of the players are shown in Fig. \ref{tandh} and the obstacles are indicated by shadows. In this environment, the agent learns to avoid the obstacles and reach one of the goal positions. In contrast, the opponent tries to grab the agent within the allowed range (right half of the environment). Therefore, the agent first needs to cross the obstacles to reach the right half of the scenario, and then get to the goal while avoiding the opponent. 

There are 5 candidate actions for each player: $ \left( up, down, left, right, standby\right) $. Once the agent reaches the goal or is caught by the opponent within permitted steps, the game is ended and the scenario will be reset.

\subsection{Experimental setup}
We evaluate the performance of HAMXCS comparing with the QbRL algorithms, including Minimax-Q \cite{littman1994markov}, Minimax-Q($\lambda$)\cite{Banerjee2001Fast}, Minimax-SARSA($\lambda$)\cite{Banerjee2001Fast}, NSCP \cite{weinberg2004best}, and HAMMQ \cite{Bianchi2014Heuristically}, and NNbRL algorithms, including DQN \cite{mnih2015human}, A3C \cite{mnih2016asynchronous}, MADDPG\cite{DBLP:journals/corr/LoweWTHAM17}, PPO\cite{schulman2017proximal}, and DPPO\cite{heess2017emergence}. Note that the best response solver of NSCP has similar ideas to the opponent modelling mechanism in \cite{uther1997adversarial}. Both of them construct the opponent model by counting the opponent's action execution frequencies. In order to compare the efficiency of using heuristics, HAMXCS and HAMMQ use the same heuristic policy during online learning.

In the following experiments, the positions of the players constitute the state or situation representation. In the two environments, we ran a sequence of 50 sessions, each of which consisted of 3000 matches. Moreover, there were 10 games in each match and the maximum steps for each player were 50.

For QbRL algorithms, the parameter settings were almost the same as the original work. The learning rate $\alpha$ was initialized to 1.0 and decayed at a rate of 0.9999954 for each iteration. The exploration rate was 0.2 and the discount factor $\gamma =0.9$. Besides, the Q-values of all $ \left(s,a,o\right)  $ tuples and the value function $ V\left(s \right)  $ of all states were initialized to 0. Moreover, the heuristic function parameters used in HAMMQ were the same as in the literature \cite{Bianchi2014Heuristically}.

For NNbRL algorithms, we adopted a similar network structure, each of which had 2 hidden layers with 200 units. See Appendix \ref{appendix} for detailed parameters. For comparison, the state representation of these algorithms were the same with HAMXCS, which is detailed in Section \ref{exp in hexcer} and Section \ref{exp in th}.

For HAMXCS, the parameters were initialized as shown in Table \ref{Initial parameters} in Appendix \ref{appendix}, most of which were identical to the original XCS.

\subsubsection{Experiments in Hexcer environment}
\label{exp in hexcer}
In this environment, we evaluated the performance of the proposed HAMXCS compared with the above RL algorithms, in which the agents were confronted with a HAMMQ opponent. The opponent's heuristic policy was symmetric with the HAMXCS as shown in Fig. \ref{hexcer}. Once a player scored a goal, it would receive a reward of $ r=100 $, otherwise $ r=0 $. 

We encoded the grids in Hexcer from left to right and from top to bottom, which was consistent with the compared QbRL algorithms. For HAMXCS and the NNbRL algorithms, we represented the situation by the players' binary coded state. Take the initial state in Fig. \ref{hexcer} as an example, the environmental state was $ \left( 23, 29\right)  $, which was composed of the grid number where the players were located. Correspondingly, the situation representation of HAMXCS was described as 110011 101111, where the first 6 digits indicated the agent's position and the last 6 digits presented the location of the opponent. In order to compare with the QbRL algorithms, we only leveraged the grid number to indicate the situation without containing other environmental knowledge.

\subsubsection{Experiments in thief-and-hunter scenario}
\label{exp in th}

Compared with Hexcer, the thief-and-hunter scenario is more challenging for the learning agents. Specifically, the agent has to learn to cross the obstacles to the right half of the environment, and then it must reach the goal without being caught. In this environment, the agents were confronted with a Minimax-Q opponent. If the agent was located in the left part of the environment, it received a penalty $ r=-10 $. Besides, the agent obtained a reward $ r=100 $, if it successfully got to the goal without being caught. Otherwise, if the agent was grabbed, the opponent was rewarded 100 as the reinforcement and the agent received a -100 punishment. 

In order to speed up the learning process, we designed 2 heuristic policies:
\begin{enumerate}
	\item [$\bullet$] The first was to guide the agent to choose actions approached the goal (measured in Manhattan distance).
	\item [$\bullet$] The second was to choose actions away from the opponent when the agent was located in the right half of the environment.
\end{enumerate}
We conducted experiments using the action selection strategies described in Section \ref{action selection} and compared the results.

We used the binary coded coordination of the players to represent the state of HAMXCS and NNbRL agents. For example, the initial state of the players in Fig. \ref{tandh} can be described as 011 000 011 110, which was composed of the coordination of the agent and the opponent, respectively. Specifically, every 6 digits represented the coordination of a player, where the first 3 digits indicated the row number and the last 3 digits were for the column.

\subsection{Results}

We first evaluate the performance of the proposed algorithm comprehensively through indicators such as average net wins, accumulated net wins, total wins, and total steps in Section \ref{result1} and \ref{result2}. Besides, we also analyze the performance differences caused by the two action selection methods in Section \ref{action selection} and different types of opponent models. Then, we detail the generalization and the interpretability of the learned classifiers in Section \ref{interpretation}. Finally, we analyze the time-consuming during the learning process in Section \ref{time analysis}. All the learned results are averaged over 50 sessions. Besides, we abbreviate Minimax-Q, Minimax-Q($\lambda$), and Minimax-SARSA($\lambda$) as MM-Q, MM-Q($\lambda$), and MM-S($\lambda$) in the figures of the results. We use HAMXCS-P to denote the usage of Pareto optimal action set in the action selection of HAMXCS.

\subsubsection{Results in Hexcer environment} \label{result1}

\begin{figure}[htbp]
	\centering
	\subfigure[Comparison of HAMXCS with QbRL algorithms.]{
		\includegraphics[scale=0.35]{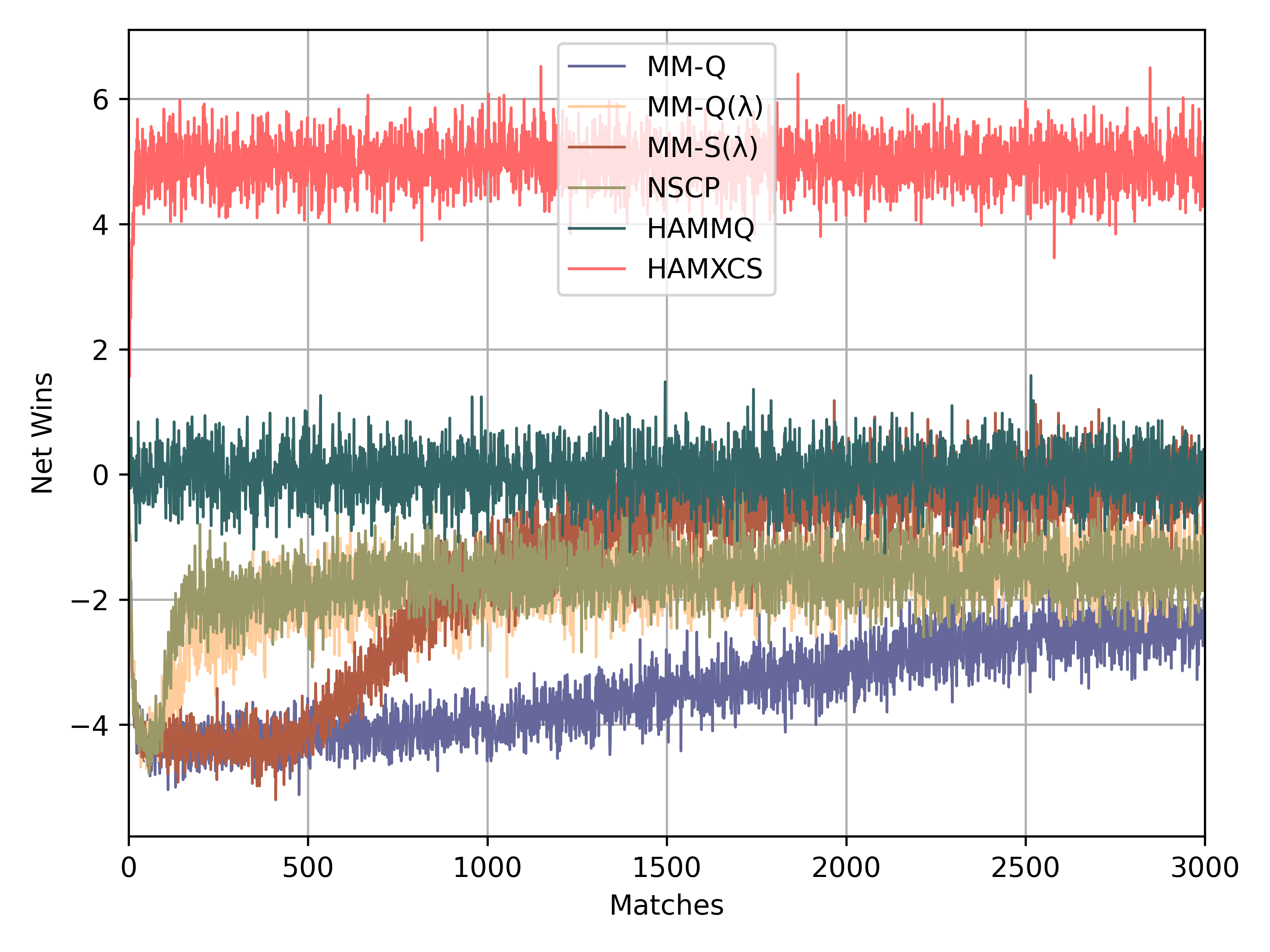}
		\label{Average goal HAMMQcomparsion with table rl in Hexcer}}
	\hfil
	\subfigure[Comparison of HAMXCS with NNbRL algorithms.]{
		\includegraphics[scale=0.35]{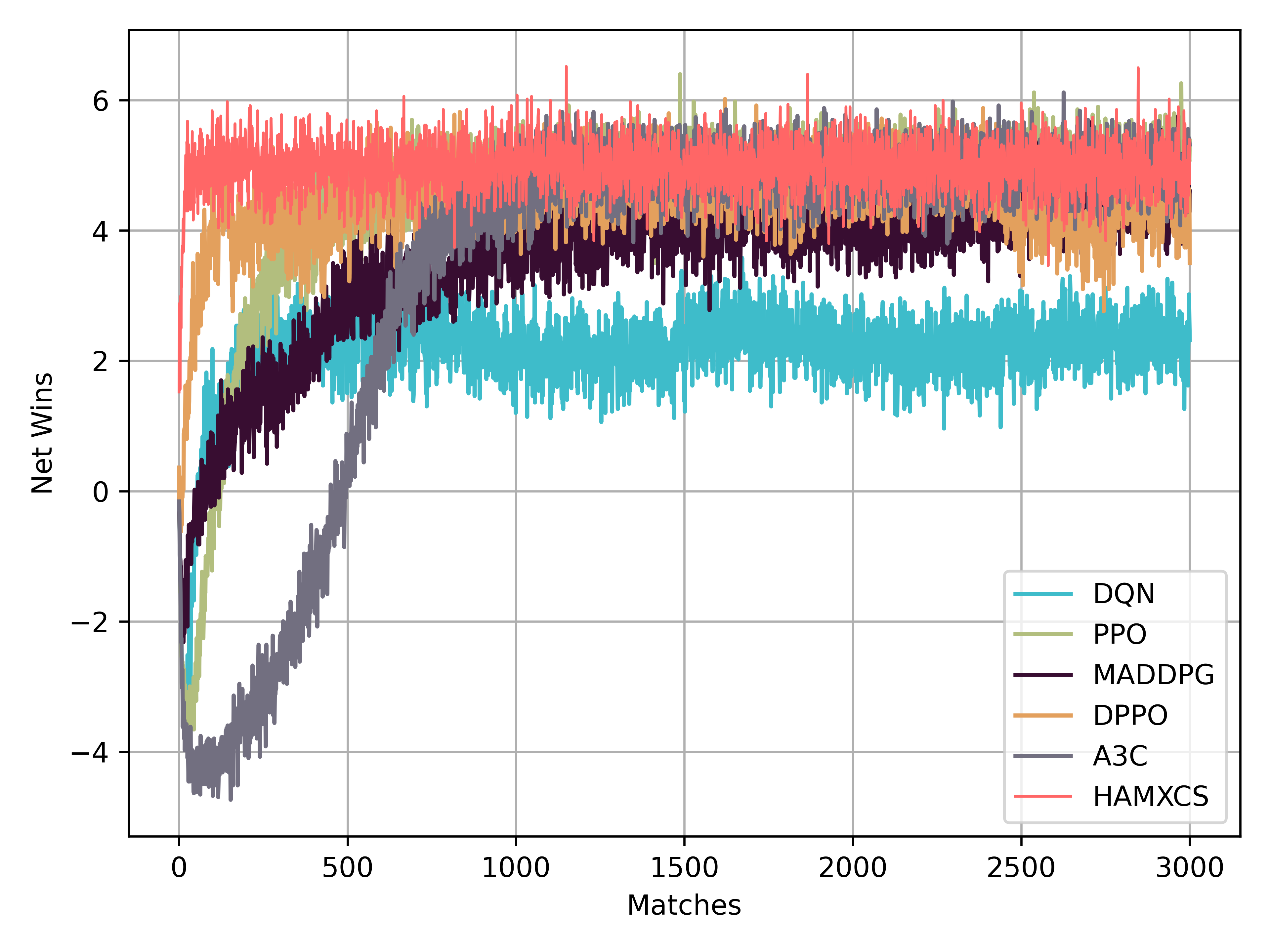}
		\label{Average goal HAMMQcomparsion with nn rl in Hexcer}}
	\caption{Averaged net wins in each match for the RL agents versus a HAMMQ opponent in Hexcer.}
	\label{Average goal HAMMQ opponent in Hexcer}
\end{figure}

\begin{figure}[htbp]
	\centering
	\subfigure[Comparison of HAMXCS with QbRL algorithms.]{
		\includegraphics[scale=0.35]{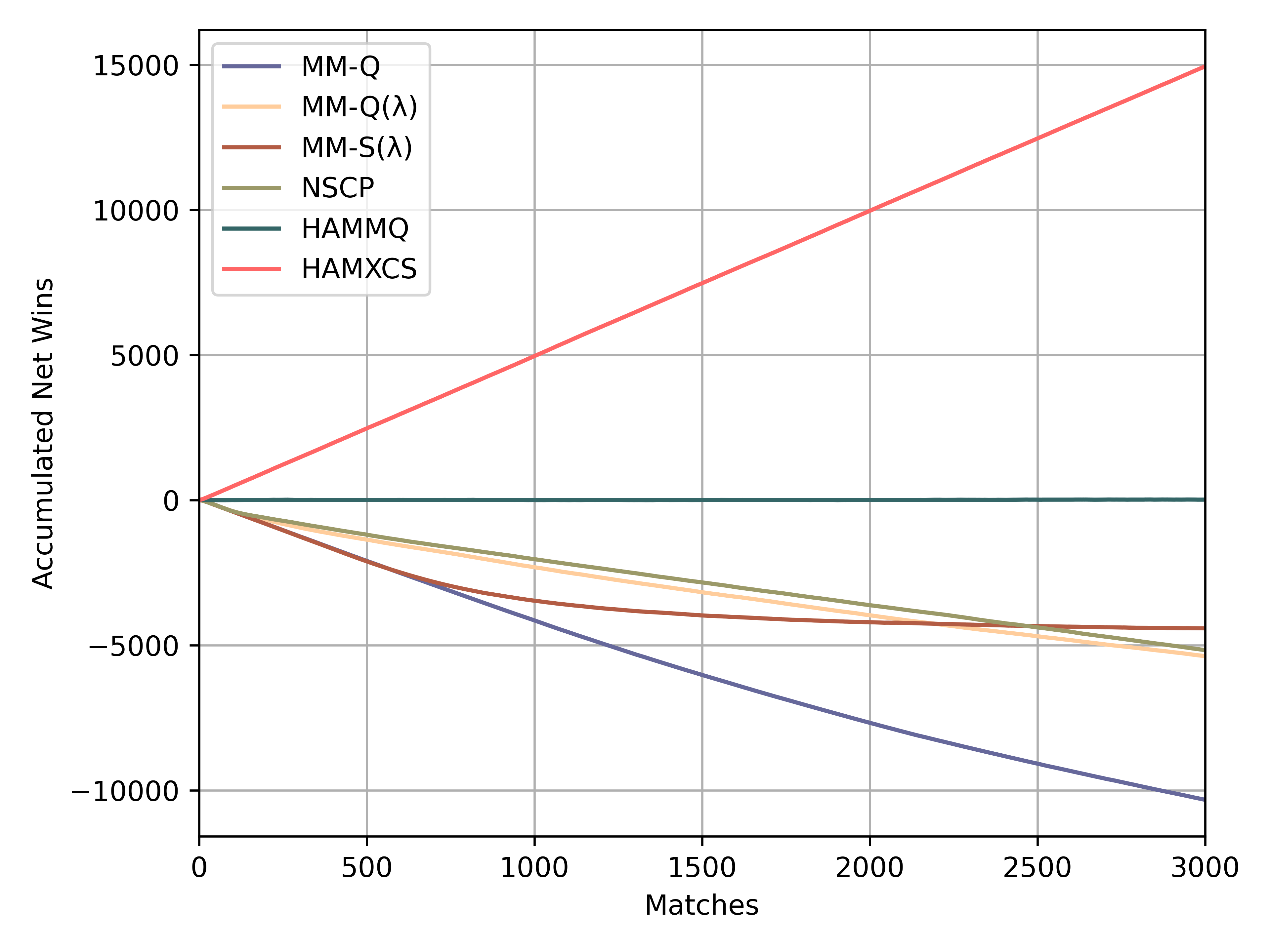}
		\label{Accumulated goal HAMMQcomparsion with table rl in Hexcer}}
	\hfil
	\subfigure[Comparison ofHAMXCS with NNbRL algorithms.]{
		\includegraphics[scale=0.35]{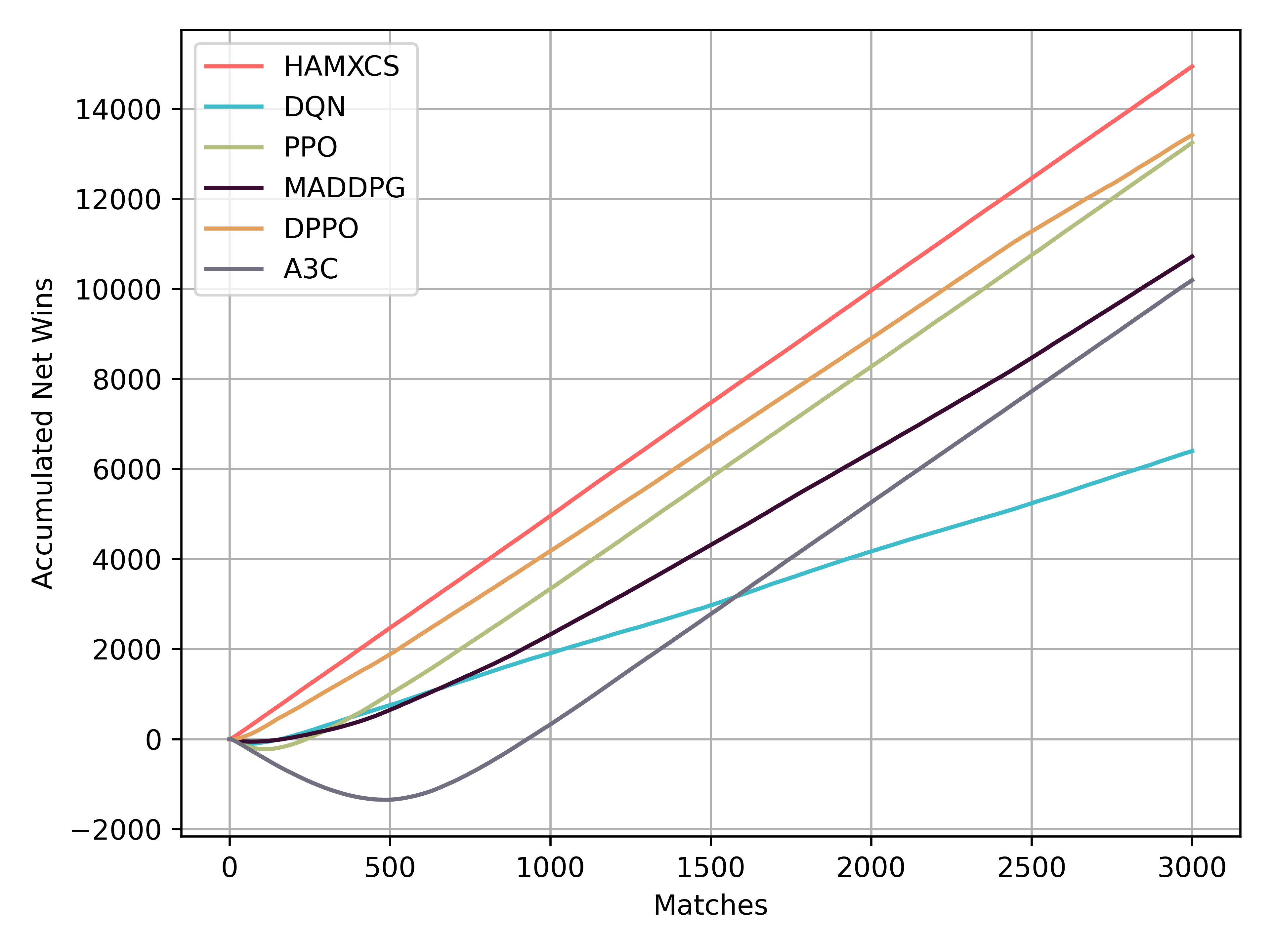}
		\label{Accumulated goal HAMMQcomparsion with nn rl in Hexcer}}
	\caption{Accumulated net wins in each match for the RL agents versus a HAMMQ opponent in Hexcer.}
	\label{Accumulated goal HAMMQ opponent in Hexcer}
\end{figure}

Fig. \ref{Average goal HAMMQcomparsion with table rl in Hexcer} and Fig. \ref{Accumulated goal HAMMQcomparsion with table rl in Hexcer} show the comparison with the QbRL agents of average net wins and accumulated net wins when against the HAMMQ opponent who used the opposite heuristic policy in Fig. \ref{hexcer}. It clearly shows that HAMXCS gets nearly 5 net goals at the beginning and maintains the advantage throughout the leaning process. Besides, HAMXCS also achieves the highest accumulated net wins. These phenomenons directly demonstrate that our approach utilizes the heuristic policy efficiently. In contrast, although HAMMQ uses the same heuristic policy as our approach, its average net wins fluctuate around 0 during the whole learning. This is because in this case, the opponent and the agent use the same learning algorithm. Besides, HAMMQ only takes advantage of the heuristic in given states without generalization. In contrast, our method can guide action selection for situations with similar feature representation.

Fig. \ref{Average goal HAMMQcomparsion with table rl in Hexcer} also shows that the average net wins of Minimax-SARSA($\lambda$) only approaches 0 at the end of learning. Compared with Minimax-Q($\lambda$) and Minimax-SARSA($\lambda$), our approach utilizes an accuracy-based eligibility trace mechanism. In other words, the more accurate the classifier, the greater the update. The learning performance indicates that the update mechanism in $ [\rm{A}]_{et} $ plays an important role in HAMXCS. Minimax-Q($\lambda$) and NSCP behaves similarly in the experiment. Moreover, NSCP maintains an opponent model by counting the opponent's action frequencies. However, it only gets better performance in the early stage of learning and is outperformed by Minimax-SARSA($\lambda$) after about 1000 matches as shown in Fig. \ref{Average goal HAMMQcomparsion with table rl in Hexcer}. In contrast to NSCP, the opponent model in our method can predict the opponent's action in situations that are feature similar to the previously visited. 

\begin{figure}[htbp]
	\centering
	\subfigure[Comparison of HAMXCS with QbRL algorithms.]{
		\includegraphics[scale=0.35]{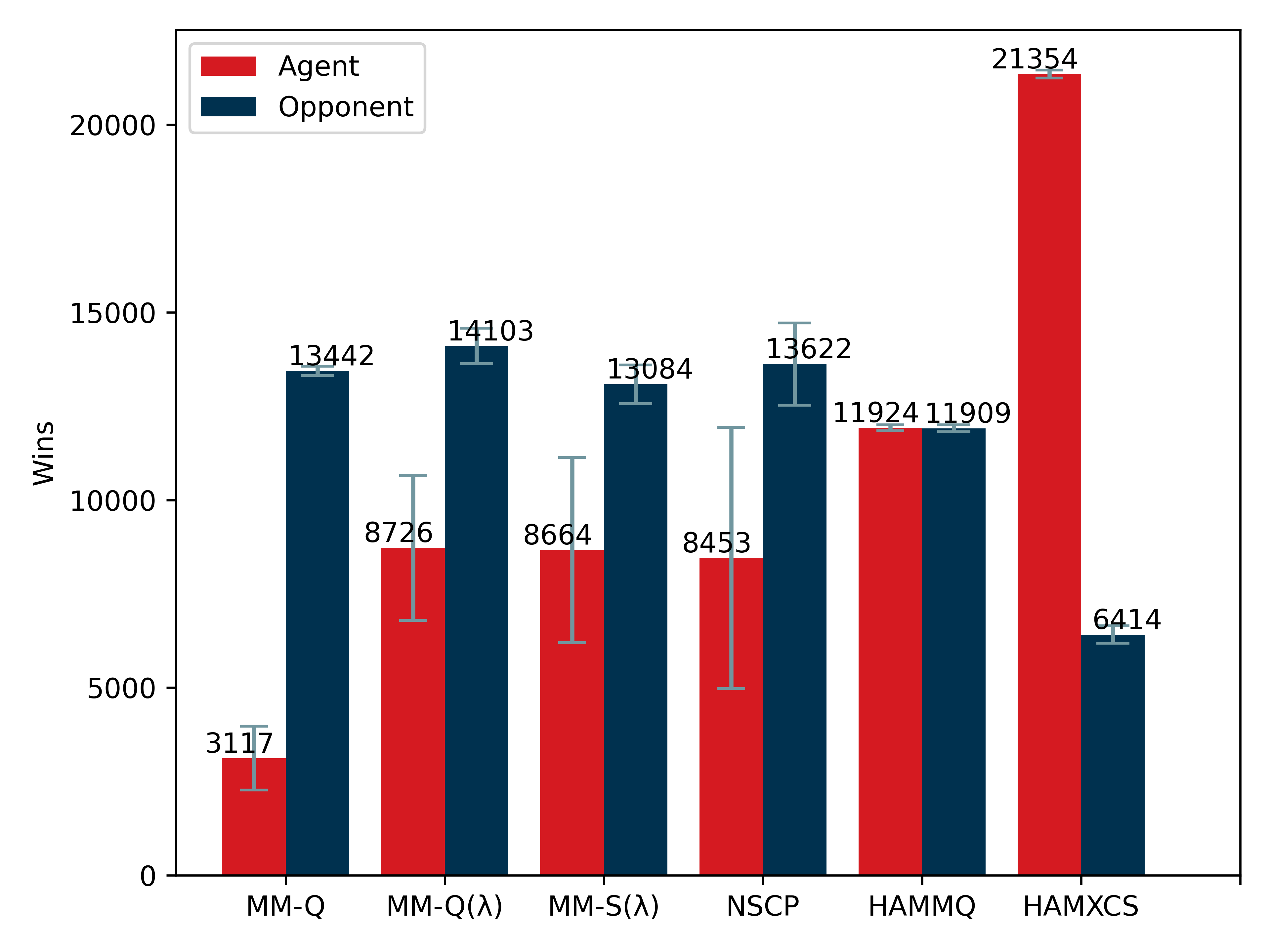}
		\label{Average cumulative number of wins HAMMQcomparsion with table rl in Hexcer}}
	\hfil
	\subfigure[Comparison of HAMXCS with NNbRL algorithms.]{
		\includegraphics[scale=0.35]{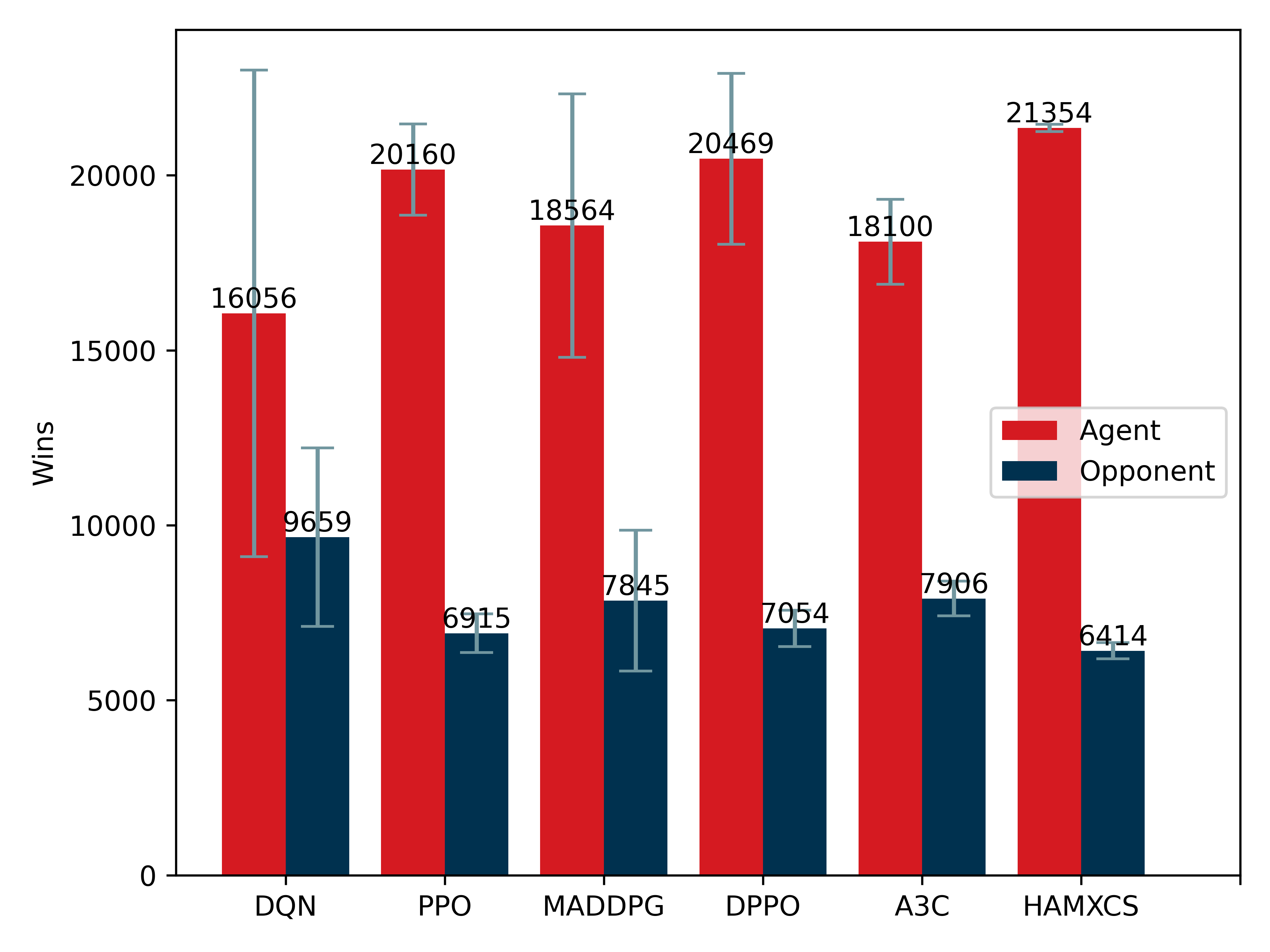}
		\label{Average cumulative number of wins HAMMQcomparsion with nn rl in Hexcer}}
	\caption{Averaged cumulative number of wins in 3000 matches of the RL agents versus a HAMMQ opponent in Hexcer. The gray lines refer to the standard deviation.}
	\label{number of wins HAMMQ opponent in Hexcer}
\end{figure}

\begin{figure}[htbp]
	\centering
	\subfigure[Comparison of HAMXCS with QbRL algorithms.]{
		\includegraphics[scale=0.35]{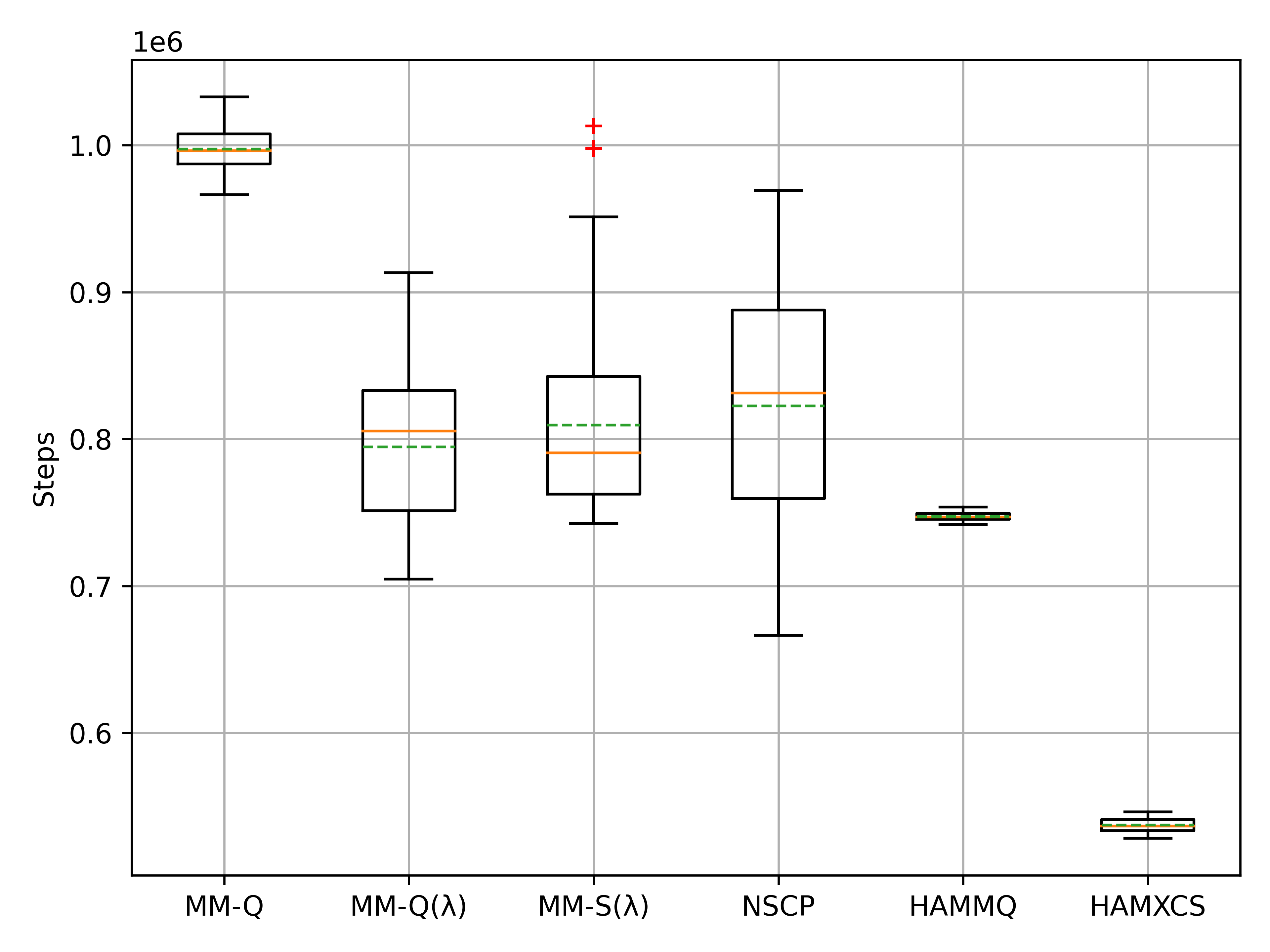}
		\label{total steps HAMMQ opponent in Hexcer HAMMQcomparsion with table rl in Hexcer}}
	\hfil
	\subfigure[Comparison of HAMXCS with NNbRL algorithms.]{
		\includegraphics[scale=0.35]{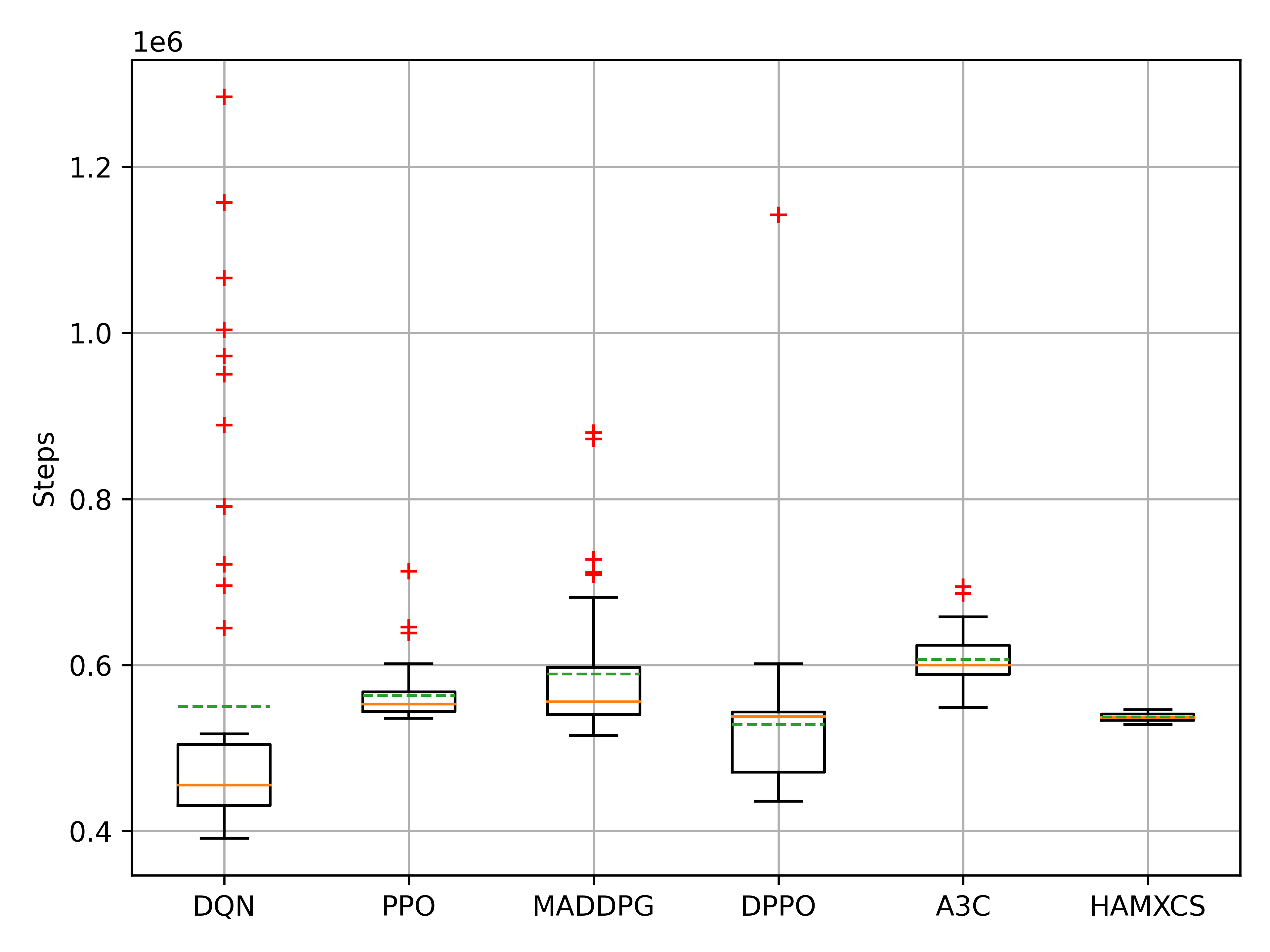}
		\label{total steps HAMMQ opponent in Hexcer HAMMQcomparsion with nn rl in Hexcer}}
	\caption{Averaged total steps in 3000 matches of the RL agents versus a HAMMQ opponent in Hexcer. The green dotted line refers to the mean.}
	\label{total steps HAMMQ opponent in Hexcer}
\end{figure}

Fig. \ref{Average cumulative number of wins HAMMQcomparsion with table rl in Hexcer} and Fig. \ref{total steps HAMMQ opponent in Hexcer HAMMQcomparsion with table rl in Hexcer} present the averaged total wins and the total steps in 3,000 matches of HAMXCS comparing with the QbRL algorithms. Please note that these two indicators can be used as a criterion to evaluate the performance of the RL agents. This is because if the games end in a small number of total steps, it means that one player's policy is dominant. Moreover, the total wins indicate which player has the upper hand. It is evident that HAMXCS obtains the most wins while using the minimum total steps compared to the QbRL algorithms. In detail, our approach uses about 20,000 steps less than HAMMQ (which performs best among QbRL algorithms), while gains about 9,500 more wins.

Next, we present the results of the NNbRL algorithms against the HAMMQ opponent. Fig. \ref{Average goal HAMMQcomparsion with nn rl in Hexcer} and Fig. \ref{Accumulated goal HAMMQcomparsion with nn rl in Hexcer} show the comparison of the average net wins of each match and the accumulated net wins. Benefiting from the efficient use of the heuristic policy and the opponent model, HAMXCS presents a noticeable advantage in the early stage of learning. Besides, the performance of DPPO is closest to that of HAMXCS in the experiments and is superior to other NNbRL algorithms. This is due to the fact that there are 4 parallel learners updating the shared model. The DPPO player only approaches the performance of HAMXCS after 1000 matches. Furthermore, MADDPG maintains a similar opponent model with HAMXCS, however, it presents an inferior performance. This indirectly illustrates the capability of HAMXCS to utilize the heuristic policy. 

Fig. \ref{Average cumulative number of wins HAMMQcomparsion with nn rl in Hexcer} and Fig. \ref{total steps HAMMQ opponent in Hexcer HAMMQcomparsion with nn rl in Hexcer} show the comparison of total wins and total steps with the NNbRL players. From the results in Fig. \ref{Average cumulative number of wins HAMMQcomparsion with nn rl in Hexcer}, it can be seen that HAMXCS delivers the best performance while its opponent obtains the least total wins. DPPO and PPO only perform slightly worse than HAMXCS. However, the total steps of the players in Fig. \ref{total steps HAMMQ opponent in Hexcer HAMMQcomparsion with nn rl in Hexcer} show that our approach distributed more evenly, which indicates that it performs more stable during online interactions.

After the experiments, HAMXCS maintains an average of 440 \textit{macroclassifiers} in the population [P], among which, only 64 \textit{macroclassifiers} has been used for action selection (i.e., $ cl.exp>0 $). In other words, our approach performs a significant learning ability and efficiency of heuristic policy application with a few frequently used classifiers.

\subsubsection{Results in thief-and-hunter scenario} \label{result2}
In this scenario, we compare the performance of the proposed approach with the NNbRL algorithms and HAMMQ which performs best among the QbRL algorithms in Hexcer. Note that HAMMQ uses identical heuristic policies as our algorithm. 

\begin{figure}[htbp]
	\centering
	\subfigure[Comparison of net wins.]{
		\includegraphics[scale=0.35]{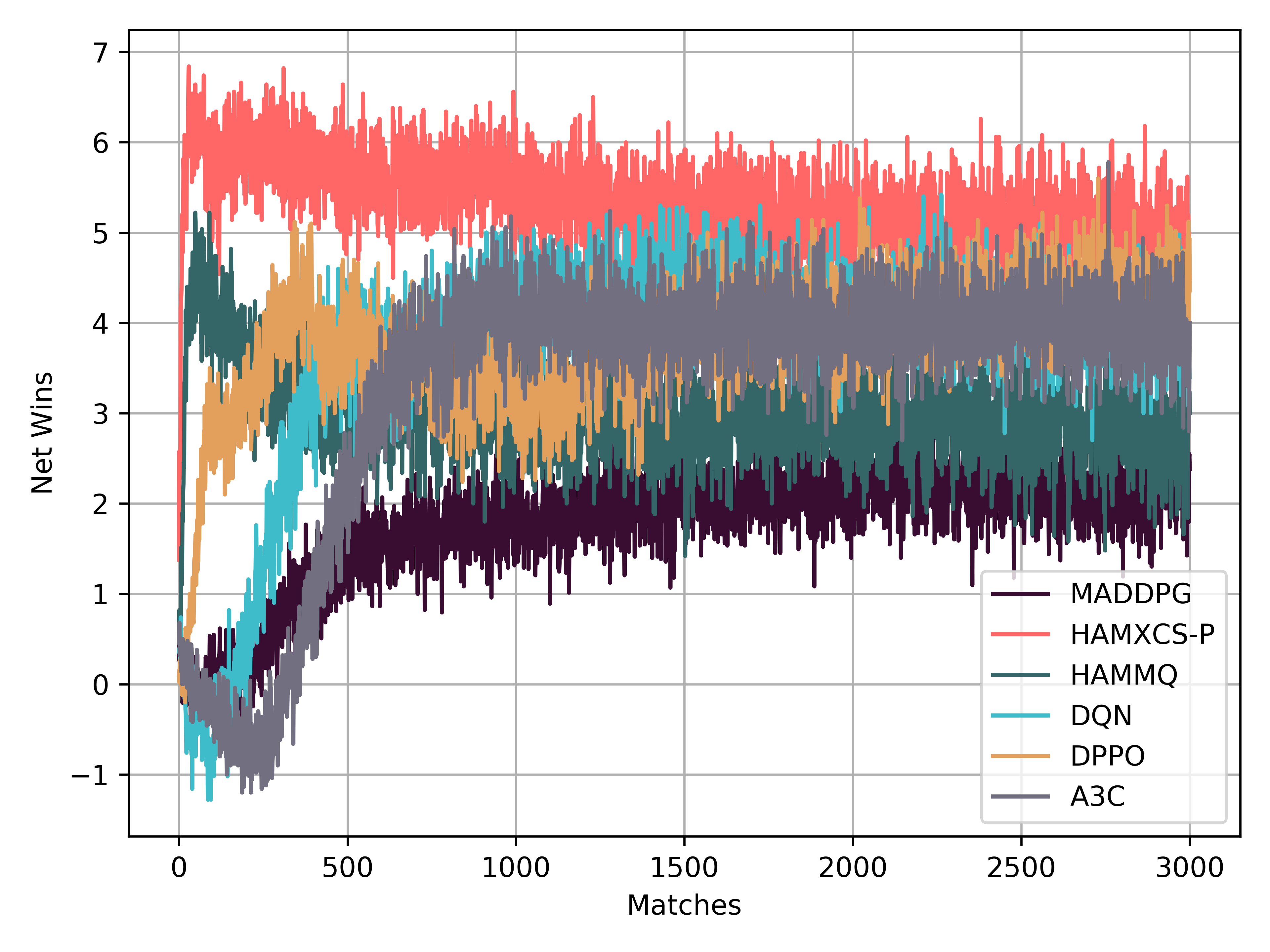}
		\label{net wins in th}}
	\hfil
	\subfigure[Comparison of accumulated net wins.]{
		\includegraphics[scale=0.35]{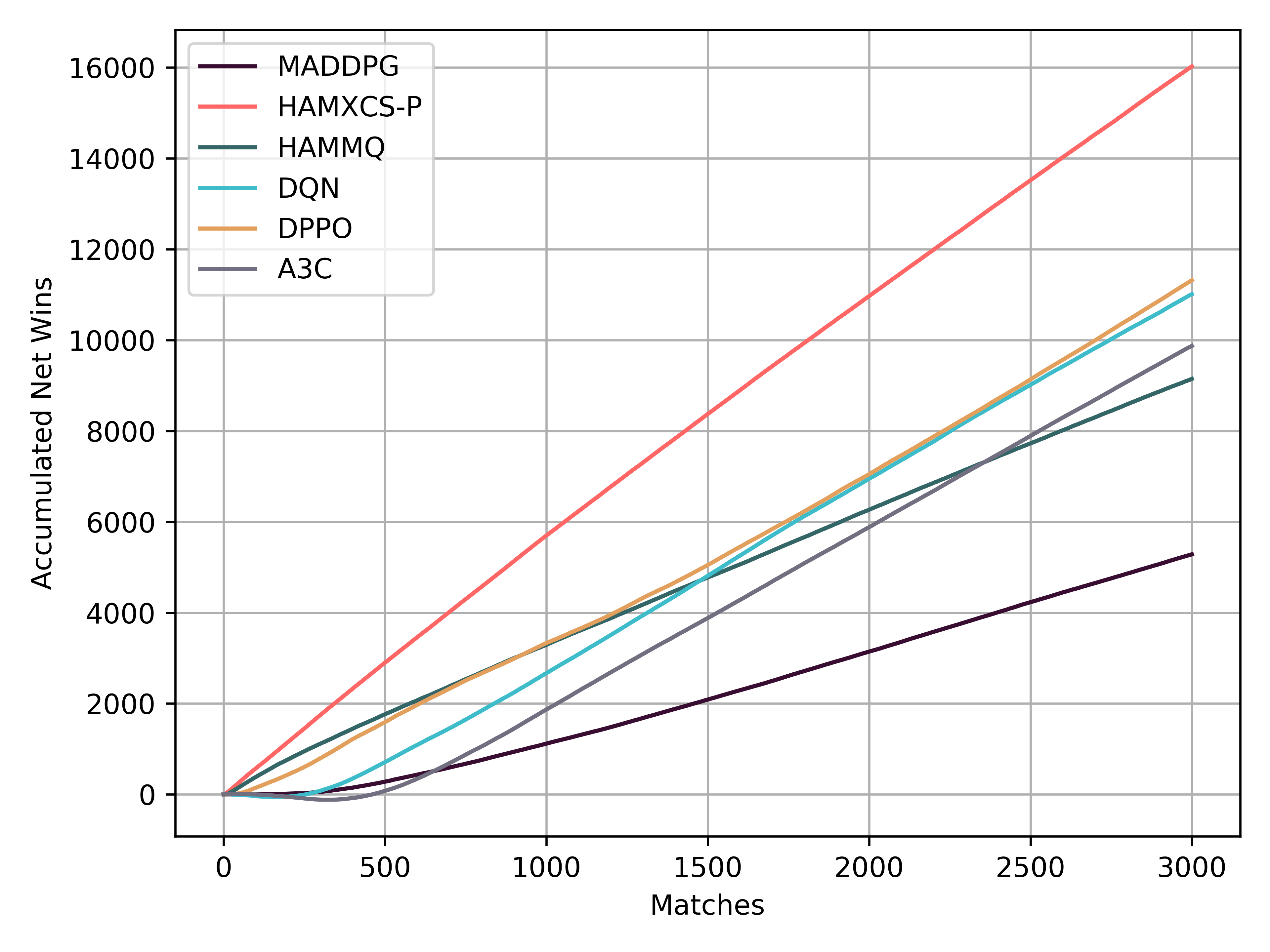}
		\label{ac wins in th}}
	\caption{Averaged and accumulated net wins in each match for the RL agents in thief-and-hunter scenario.}
	\label{goals in th}
\end{figure}

\begin{figure}[htbp]
	\centering
	\subfigure[Comparison of total wins.]{
		\includegraphics[scale=0.35]{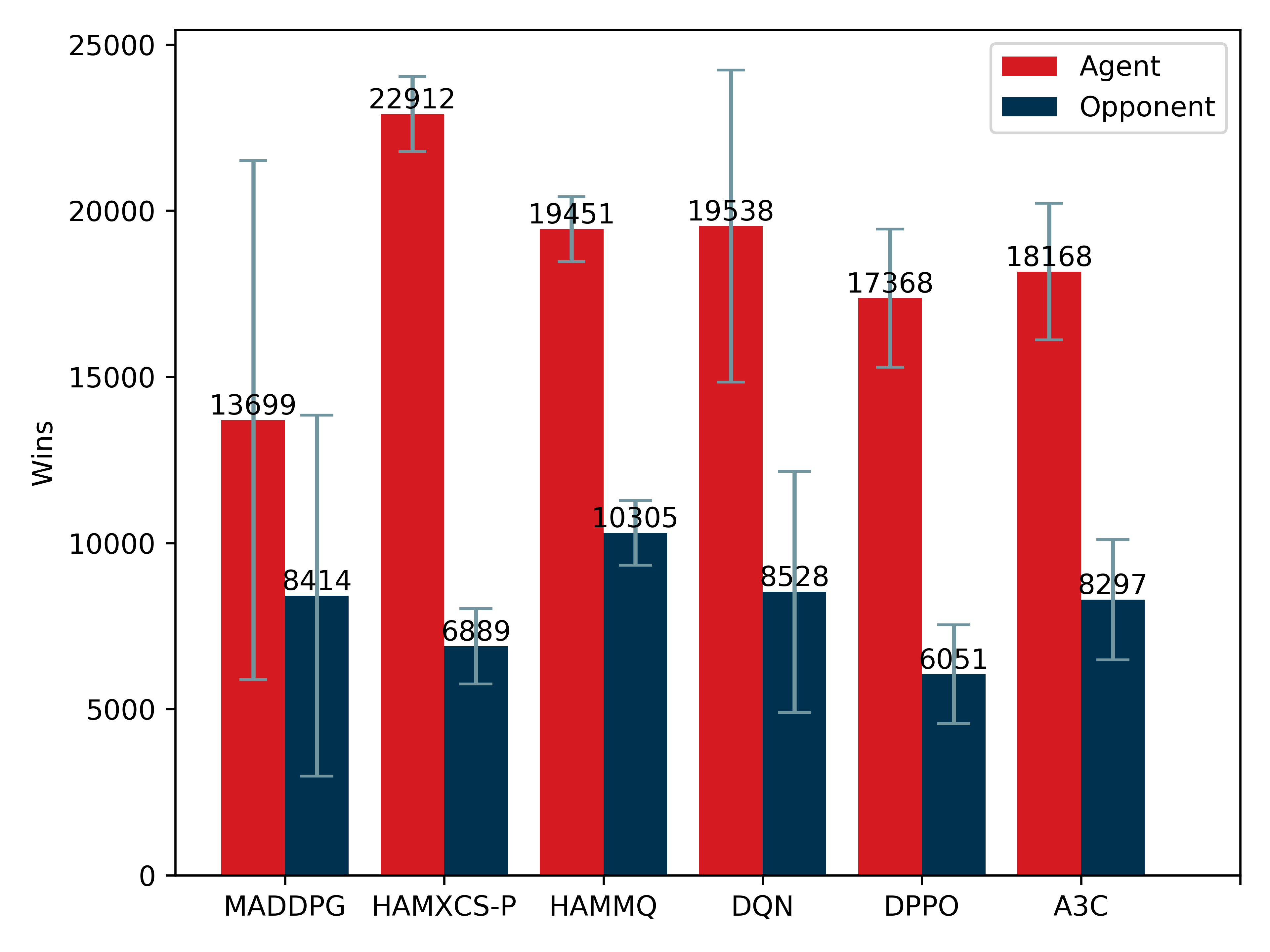}
		\label{wins in th}}
	\hfil
	\subfigure[Comparison of total steps.]{
		\includegraphics[scale=0.35]{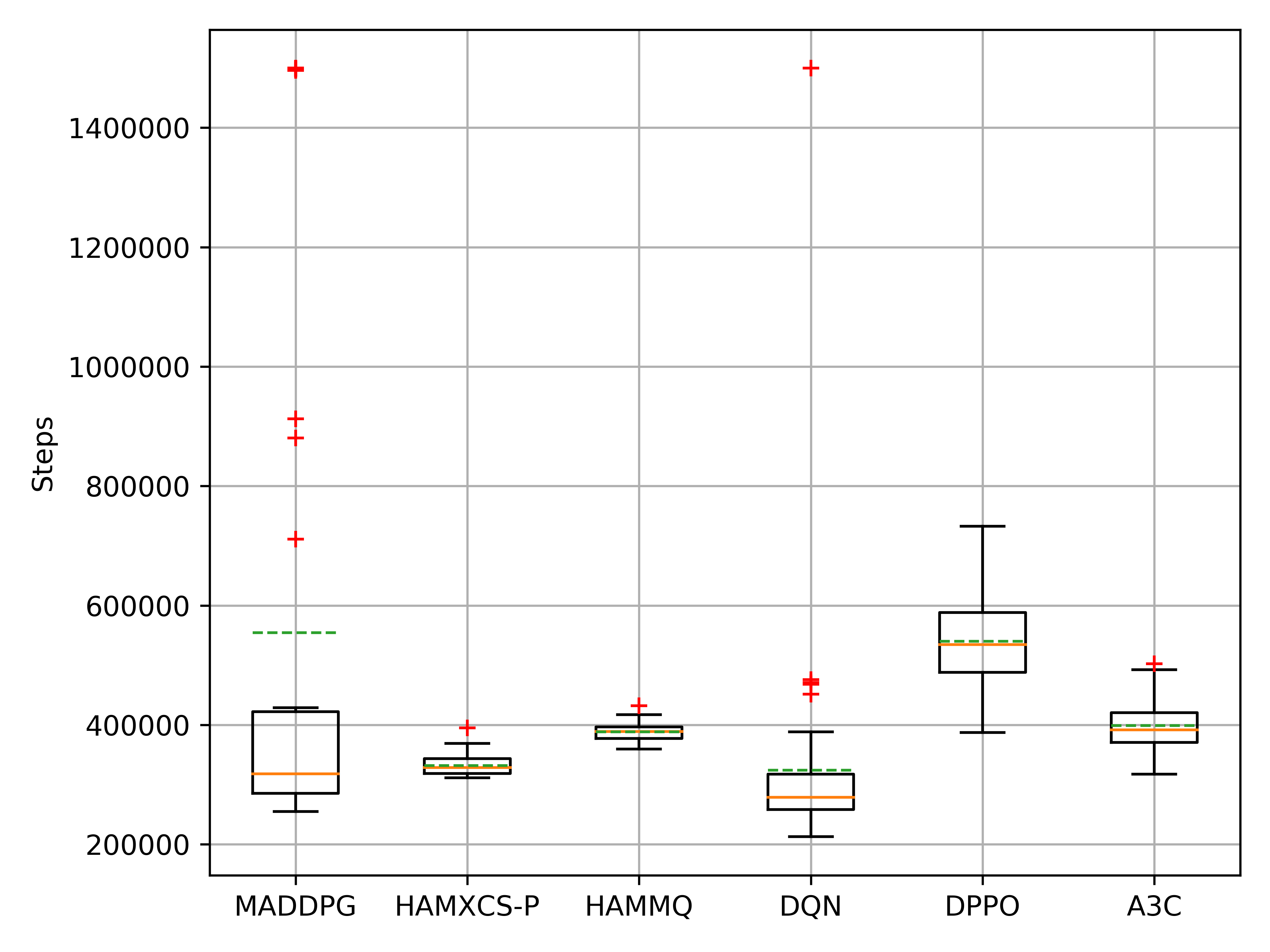}
		\label{steps in th}}
	\caption{Averaged total wins and total steps of the RL agents in thief-and-hunter scenario.}
	\label{wins and steps in th}
\end{figure}

Fig. \ref{net wins in th} and Fig. \ref{ac wins in th} show the comparison of net wins and accumulated net wins averaged over 50 sessions. It is evident that HAMXCS-P delivers significantly better performance than other algorithms. This is because our approach thoroughly considers the 2 heuristic policies of approaching the goal position and avoiding the opponent, and selects the action from the Pareto optimal action set. In contrast,  HAMMQ uses the same heuristic policies as our approach, however, it only gains the upper hand in the early stage of learning compared with the NNbRL algorithms, such as DPPO. The net wins of these two heuristic algorithms decrease gradually because the opponent is also evolving its policy. DPPO achieves the highest accumulated net wins among NNbRL algorithms because its opponent wins the least in the experiments, as shown in Fig. \ref{wins in th}. A3C and DQN show better performance in this experiment. After about 1000 matches, these two algorithms behave similarly. Similar results can be found in Fig. \ref{wins in th} and Fig. \ref{steps in th}. It shows that A3C performs more stable than DQN because the standard deviation of A3C is smaller than DQN in Fig. \ref{wins in th}. It also clearly shows that our approach performs best among the algorithms. Specifically, HAMXCS-P gets the most wins while using relatively small number of total steps. Furthermore, in this experiment, our approach maintains an average of 328 \textit{macroclassifiers} in the population [P], of which only 242 have been used for action selection.

\begin{figure}[htbp]
	\centering
	\subfigure[Comparison of net wins.]{
		\includegraphics[scale=0.35]{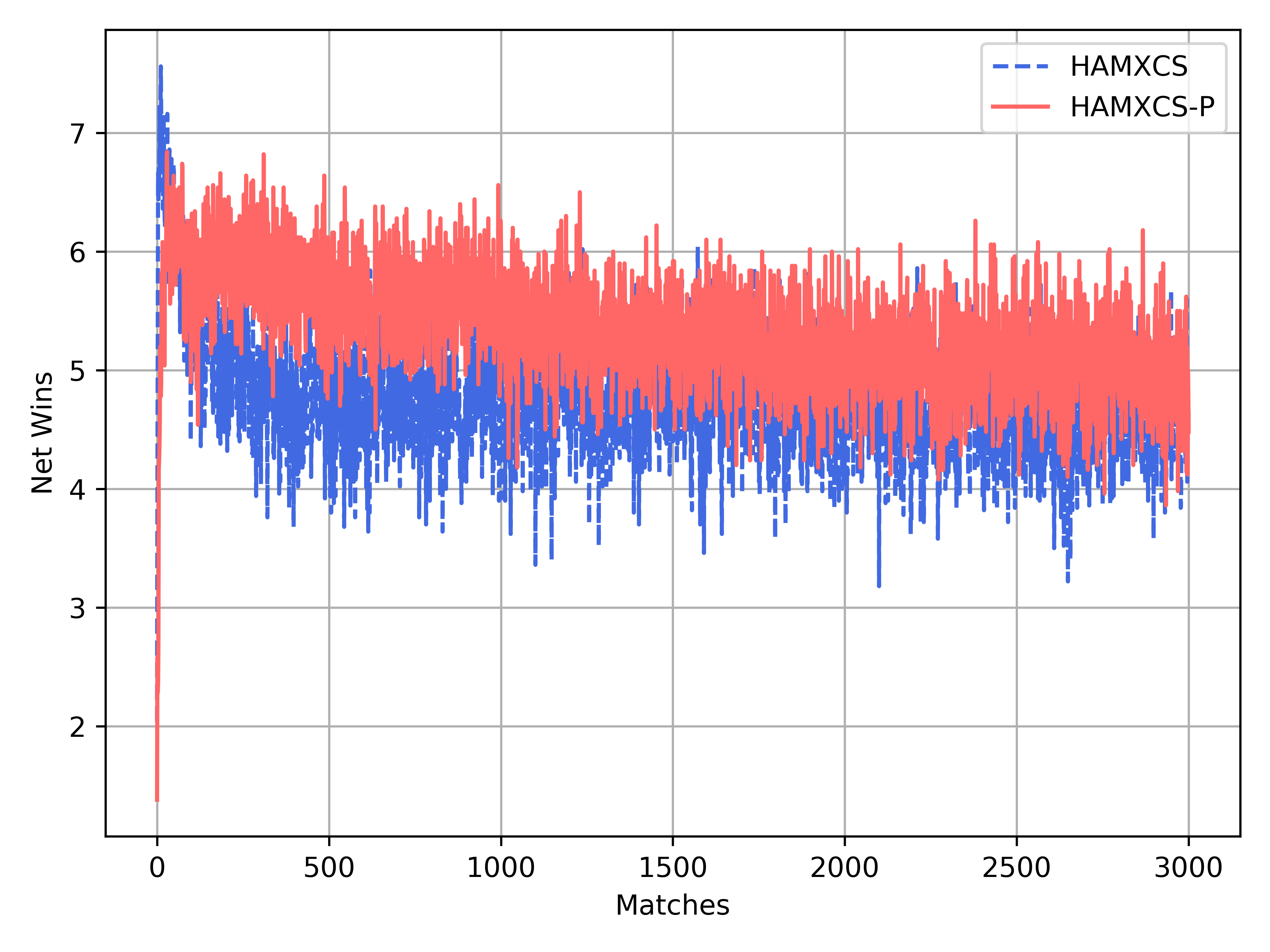}
		\label{TH_COMP_ASMSrun}}
	\hfil
	\subfigure[Comparison of accumulated net wins.]{
		\includegraphics[scale=0.35]{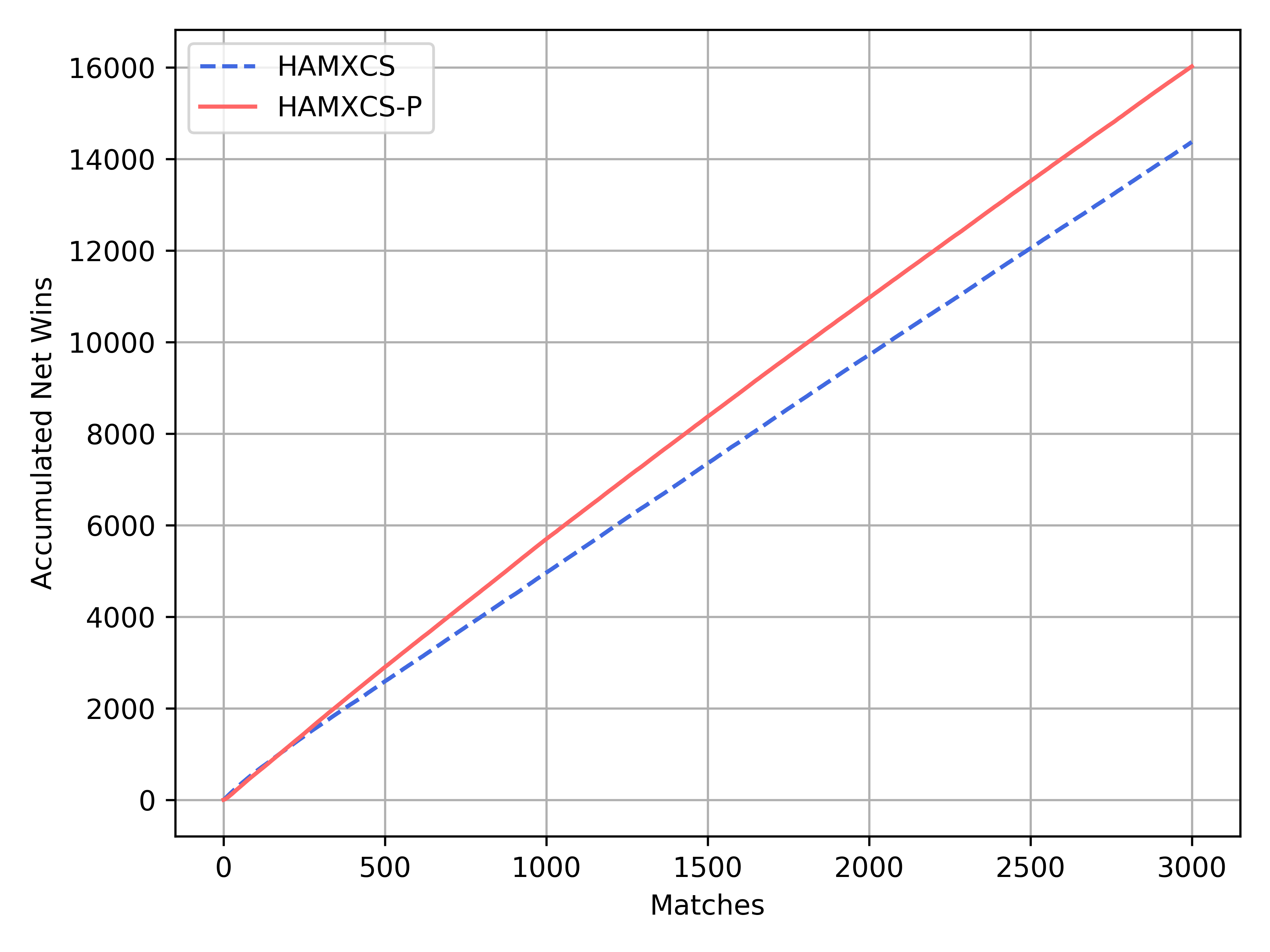}
		\label{TH_COMP_ASMSac_reward}}
	\hfil
	\subfigure[T-test of net wins.]{
		\includegraphics[scale=0.35]{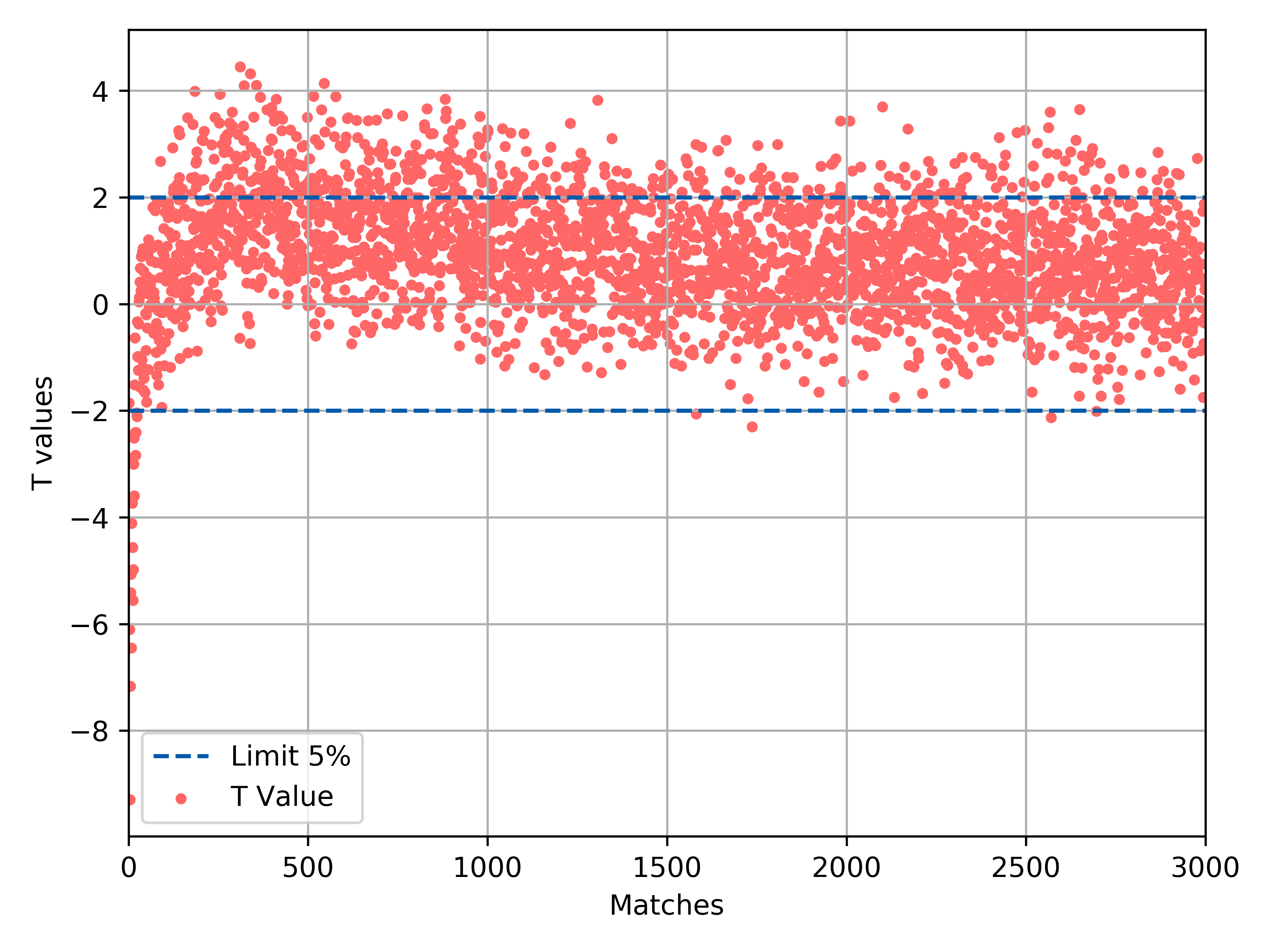}
		\label{TH_COMP_ASMSt_test}}
	\caption{Performance comparison between HAMXCS and HAMXCS-P.}
\end{figure}

Next, we evaluate the action selection methods in Section \ref{action selection}. Here, HAMXCS takes the sum of the 2 heuristic vectors (corresponding to the 2 given heuristic policies) as the final heuristic, while HAMXCS-P chooses action among Pareto optimal actions. Fig. \ref{TH_COMP_ASMSrun} and Fig. \ref{TH_COMP_ASMSac_reward} show the comparison of net wins and accumulated net wins respectively. It is evident that HAMXCS-P performs better in the experiment. Specifically, after 3000 matches, HAMXCS-P obtains about 2000 more accumulated net wins than HAMXCS. This is because HAMXCS-P comprehensively considers the 2 heuristic policies and selects actions in the Pareto optimal manner. Similar comparison results can be found in Fig. \ref{TH_COMP_ASMSt_test}, which shows the T-test of averaged net wins over 50 sessions between HAMXCS and HAMXCS-P. It indicates that HAMXCS-P performs better in some games after about 100 matches with a confidence level greater than 95\%, which is consistent with the results in Fig.\ref{TH_COMP_ASMSrun}. Although HAMXCS-P behaves better in the experiments, we note that HAMXCS-P consumes more time than HAMXCS, which will be detailed in Section \ref{time analysis}.

\begin{figure}[htbp]
	\centering
	\subfigure[Comparison of accumulated net wins.]{
		\includegraphics[scale=0.35]{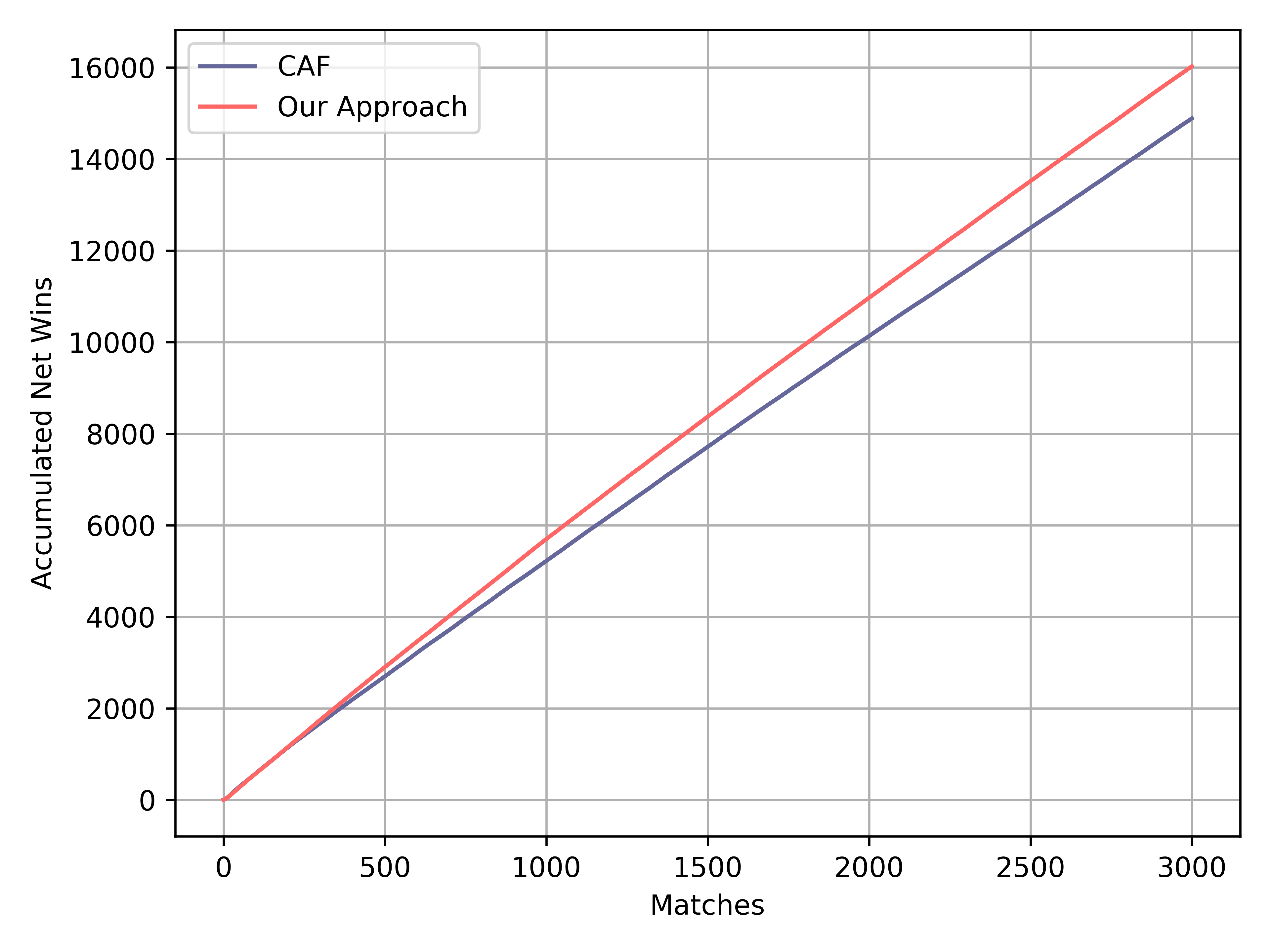}
		\label{TH_COMP_OMSac_reward}}
	\hfil
	\subfigure[T-test of accumulated net wins.]{
		\includegraphics[scale=0.35]{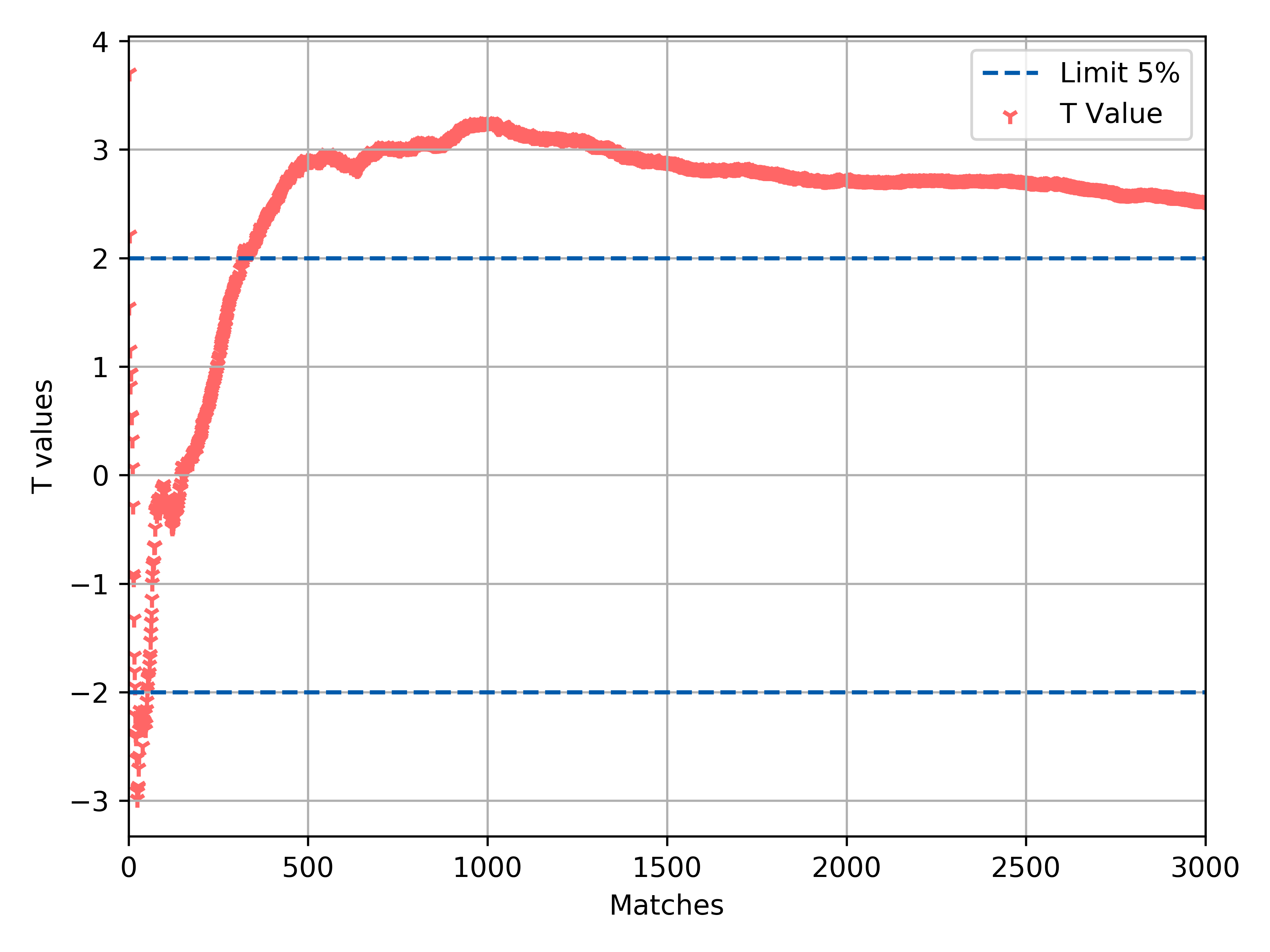}
		\label{TH_COMP_OMSt_test}}
	\caption{Performance comparison of HAMXCS-P between using CAF and our opponent model.}
\end{figure}

In order to evaluate the efficiency of the opponent model in Section \ref{opponent model}, we compare the performance of HAMXCS-P when using our opponent model and when just counting the opponent's action frequencies (CAF). Fig. \ref{TH_COMP_OMSac_reward} shows the accumulated net wins differences caused only by different opponent models. It is evident that our approach delivers a noticeable better performance. Specifically, after 3000 matches, our approach gets 1000 more net wins than CAF. Similar results can be seen in Fig. \ref{TH_COMP_OMSt_test}, which demonstrates the T-test of accumulated net wins of HAMXCS-P between using CAF and our opponent model. It clearly shows that after about 300 matches, our approach performs a significant advantage. This is due to the fact that our approach can predict the opponent's behavior in states with similar features.

Summarizing the results in both environments, it is evident that HAMXCS(-P) delivers significantly better performance than QbRL algorithms and performs similar learning ability or even better than some NNbRL algorithms. This is due to the fact that HAMXCS makes efficient use of the heuristic policies, on the other hand, it also benefits from the classifiers update mechanism proposed in this paper. It is worth highlighting that  HAMXCS shows a significant learning performance while using a relatively small number of classifiers.

\subsubsection{Interpretation of the learned classifiers} \label{interpretation}
In this section, we present some learned classifiers after the experiments and detail its interpretability and generalization. In the following results, the accuracy of the prediction vector and heuristic vector retains one decimal place, and the error and fitness retain two decimal places. Besides, the prediction value corresponding to the action advocated by the classifier is highlighted in bold.

\begin{table}[htbp]
	\centering
	\caption{Examples of the learned classifiers in Hexcer.}
	\label{classifiers in hexcer}
	\resizebox{\linewidth}{!}
	{
		\begin{tabular}{cccccccc}	
			\hline\noalign{\smallskip}
			No.&Condition & Action & Prediction vector &Heuristic vector &Fitness & Error &  Experience \\
			\noalign{\smallskip}\hline\noalign{\smallskip}
			1&$ \#101100\#11\#\#  $&  Right&[0.0 0.0 5.0 $ \bm{8.6} $ 2.6 0.0 17.7] & [22.7 0.00 46.9 31.8 64.8 89.2 10.0]& 1.00 & 0.09 & 39\\
			2&$ \#\#100\#\#\#\#\#\#\# $ &  Right&[32.7 42.0 41.7 $ \bm{37.5} $ 37.4 35.7 32.9] & [20.1 20.4 10.0 10.8 22.3 28.8 10.0]& 0.95 & 0.01 & 357\\
			3&$ \#\#1111\#\#\#\#\#\# $ &  Right& [26.0 27.3 23.8 $ \bm{25.0} $ 24.1 25.8 25.1]& [20.3 41.9 24.1 23.6 23.3 20.1 18.0]& 1.00 & 0.04 & 796\\
			4&$ 0\#\#111\#\#\#\#0\# $ &  Right & [37.5 40.2 43.4 $ \bm{37.1} $ 40.0 40.4 35.0]& [20.4 17.9 32.5 34.3 33.9 19.4 22.8]& 1.00 & 0.03 & 208\\
			\noalign{\smallskip}\hline
		\end{tabular}
	}
\end{table}

Table \ref{classifiers in hexcer} shows the learned classifiers in Hexcer against a HAMMQ opponent. It clearly shows that HAMXCS learns the most accurate and general classifiers, such as the 2nd, 3rd, and 4th classifiers. Specifically, the fitness of these classifiers approaches 1.0 and their error is close to 0.0. Besides, the wildcard \# occupies most of the bits of the classifiers. As a result, the classifier whose condition matches given situations can be used for action selection, i.e., a classifier can be used in multiple situations. Furthermore, HAMXCS also learns the interpretable classifiers. Take the first classifier in Table \ref{classifiers in hexcer} as an example, after decoding the condition, it can be interpreted as: if the agent is located in the 22nd grid and the opponent is in the 12$\sim$15th or 28$\sim$31st grid, the agent is advocated choose action $ Right $. It clearly demonstrates that HAMXCS not only can learn classifiers with generalization ability, but also classifiers that humans can understand. 

It is important to emphasize that the situation representation of HAMXCS in Hexcer is not optimal for learning and interpretation. For example, we can take the ball possession and the relative positions of the players into situation representation. Besides, in order to express the commonness of the states, the players' coordination can also be considered. However, for comparison with the QbRL algorithms, we choose the representation method in this paper.

\begin{table}[htbp]
	\centering
	\caption{Examples of the learned classifiers using HAMXCS in thief-and-hunter scenario.}
	\label{classifiers in th}
	\resizebox{\linewidth}{!}
	{
		\begin{tabular}{cccccccc}	
			\hline\noalign{\smallskip}
			No.&Condition & Action & Prediction vector &Summed heuristic vector &Fitness & Error &  Experience \\
			\noalign{\smallskip}\hline\noalign{\smallskip}
			1&$010110000101  $&  Down &[0.0 0.0 $ \bm{0.7} $ 0.0 0.0] & [0.0 10.0 11.6 14.6 0.0]& 1.00 & 0.02 & 59\\
			2&$ 001100100110 $ &  Up& [$ \bm{0.0} $ 0.0 0.0 0.0 0.0]&  [13.5 14.0 14.1 15.2 18.9]& 1.00 & 0.00 & 22\\
			3&$\#1101\#100110$ &  Up&[-$ \bm{9.8} $ -10 -9.9 -9.9 -9.8] &[19.8 20.0 19.9 19.9 19.8]& 1.00 & 0.14 & 134\\
			4&$ 0\#\#00\#0001\#0 $ &  Right & [-10.0 -$ \bm{10.0} $ -10.0 -10.0 -9.9]& [19.9 19.9 20.0 20.0 19.8]& 1.00 & 0.07 & 258\\
			5&$ 0\#\#00\#\#\#\#\#\#\# $ &  Right & [-8.7 -$ \bm{8.5}$ -9.0 -8.7 -8.7]& [19.4 35.1 19.5 19.5 28.5]& 0.92& 0.12 & 1364\\
			\noalign{\smallskip}\hline
		\end{tabular}
	}
\end{table}

Table \ref{classifiers in th} gives several examples of the learned classifiers using HAMXCS in thief-and-hunter scenario. In this environment, our approach makes efficient and reasonable use of the given heuristic policies. In detail, the 1st classifier is interpreted as: if the HAMXCS agent is in row 2 and column 6, and the opponent is located in row 0 and column 5, the agent should choose action $ Down $. This is reasonable because although the agent is close to the upper goal, the opponent is near the goal position. Therefore, in order to avoid being caught by the opponent, the agent should approach the lower goal. In contrast, we can interpret the 2nd classifier as: if the agent is in row 1 and column 4, and the opponent is in row 4 and column 6, the agent should choose action $ Up $. In this case, the agent and the opponent are located in the upper part and the lower part of the environment separately, thus the agent should approach the upper goal. Similar results can be found in the 3rd classifier, where the agent is on the symmetry axis of the scenario (row 3 and column 4) and the opponent is in the same position as the 2nd classifier (lower part of the scenario). As a result, the agent is advised to choose action $ Up $ to avoid the opponent.

Moreover, HAMXCS also learns classifiers with generalization capability, such as the 4th and the 5th classifiers. Both classifiers advocate choosing action $ Right $. Furthermore, the 5th classifier can be understood as: if the row number of the agent is less than 4 and the column number is less than 2, the agent can choose action $ Right $.

After the above analysis, it is clear that HAMXCS can learn classifiers with generalization ability. Besides, it also learns interpretable classifiers for specific situations or parts of the environment. These results consistently demonstrate HAMXCS can efficiently utilize the rough heuristic policies and improve its own policy on this basis. For example, the 1st classifier in Table \ref{classifiers in th} shows that HAMXCS has learned to weight the pros and cons between reaching the goal position and avoiding the opponent. In addition, we can also get the quantitative description of a classifier based on its fitness and error.

\subsubsection{Time-consuming analysis} \label{time analysis}
In this section, we analyze the time required for learning in the experiments. Due to the space limitation, we only show the time-consuming in the thief-and-hunter scenario. The experiments were implemented in Python in a Microsoft Windows computer with 16-Core Intel i9-9900K CPU, 64Gb of RAM and GeForce GTX 2080Ti GPU. The neural networks were all implemented in TensorFlow 1.14.0. Table \ref{learning time} shows the averaged time required over 50 sessions of the algorithms in the thief-and-hunter scenario. 

\begin{table}[htbp]
	\centering
	\caption{Time required of each session in thief-and-hunter scenario.}
	\label{learning time} 
	\begin{tabular}{cc}	
		\hline\noalign{\smallskip}
		Algorithms & Time required($ s $) ($ \pm $ std)\\ 
		\noalign{\smallskip}\hline\noalign{\smallskip}
		HAMMQ & 226.3 ($ \pm $20.0) \\
		DQN & 453.5 ($ \pm $284.2) \\
		A3C& 876.3 ($ \pm $114.5)\\
		A3C (GPU)& 569.9 ($ \pm $55.1)\\
		PPO& 744.0 ($ \pm $165.8) \\
		MADDPG& 434.4 ($ \pm $303.9) \\
		DPPO (GPU)& 1154.9 ($ \pm $140.9) \\
		\noalign{\smallskip}\hline\noalign{\smallskip}
		HAMXCS & 1250.5 ($ \pm $146.5) \\
		HAMXCS-P & 2376.8 ($ \pm $133.2) \\
		\noalign{\smallskip}\hline
	\end{tabular}
\end{table}

It clearly shows that HAMXCS and HAMXCS-P require much more time to get excellent performance. On the one hand, it depends on the update mechanism of our algorithm. On the other hand, we note that the running time is also greatly affected by the population size because it takes more time to evolve more classifiers. Besides, it is evident that HAMXCS-P takes more time than HAMXCS. This is because HAXMCS-P needs to get the Pareto optimal actions. Therefore, with the number of heuristic policies increases, it is vital to choose the approximated approaches to obtain Pareto optimal actions.

\section{Conclusion and future works}
\label{conclusion}

This paper focuses on the efficient use of heuristic policies and proposes the HAMXCS to solve competitive Markov games. The heuristics are incorporated into the classifier representation and action selection to guide policy learning. Moreover, we present the upper bound of the EAP error resulted by introducing the heuristics. Besides, the opponent model is constructed during online interactions while the corresponding opponent's behavior predictions are used for action selection and classifiers evolution. Benefiting from the generalized representation of the classifiers, the opponent model and heuristic component can guide policy learning in states with similar features. Furthermore, the accuracy-based eligibility trace mechanism further speeds up the learning process.

Extensive simulations demonstrate the efficient use of the heuristics while the learned classifiers are generalized and interpretable. However, one limitation of our algorithm is that it requires more time in updating classifiers, especially when utilizing the Pareto optimal action selection strategy. Moreover, the learning time also depends on the size of the population. As future work, we will further improve the learning efficiency of the proposed approach by finding more proper action exploration strategies. On the other hand, it's worthwhile investigating how to extend HAMXCS to continuous scenarios with multiple players.

\section*{acknowledgement}
This work is supported by the National Nature Science Foundation of China under Grant No.61906203 and the 13th Five-Year equipment pre-research sharing technology project No.41412030301.

\section*{References}
\bibliography{HAMXCS}   

\begin{thebibliography}{10}
\expandafter\ifx\csname url\endcsname\relax
  \def\url#1{\texttt{#1}}\fi
\expandafter\ifx\csname urlprefix\endcsname\relax\def\urlprefix{URL }\fi
\expandafter\ifx\csname href\endcsname\relax
  \def\href#1#2{#2} \def\path#1{#1}\fi

\bibitem{sutton2018reinforcement}
R.~S. Sutton, A.~G. Barto, Reinforcement learning: An introduction, MIT press,
  2018.

\bibitem{bucsoniu2010multi}
L.~Bu{\c{s}}oniu, R.~Babu{\v{s}}ka, B.~De~Schutter, Multi-agent reinforcement
  learning: An overview, in: Innovations in multi-agent systems and
  applications-1, Springer, 2010, pp. 183--221.

\bibitem{littman1994markov}
M.~L. Littman, Markov games as a framework for multi-agent reinforcement
  learning, in: Machine learning proceedings 1994, Elsevier, 1994, pp.
  157--163.

\bibitem{bowling2002multiagent}
M.~Bowling, M.~Veloso, Multiagent learning using a variable learning rate,
  Artificial Intelligence 136~(2) (2002) 215--250.

\bibitem{watkins1992q}
C.~J. Watkins, P.~Dayan, Q-learning, Machine learning 8~(3-4) (1992) 279--292.

\bibitem{Ribeiro2002Experience}
C.~H.~C. Ribeiro, R.~Pegoraro, A.~H.~R. Costa, Experience generalization for
  concurrent reinforcement learners: the minimax-qs algorithm, in:
  International Joint Conference on Autonomous Agents \& Multiagent Systems,
  2002.

\bibitem{Banerjee2001Fast}
B.~Banerjee, S.~Sen, P.~Jing, Fast concurrent reinforcement learners, in:
  International Joint Conference on Artificial Intelligence, 2001.

\bibitem{Bianchi2007Heuristic}
R.~A.~C. Bianchi, C.~H.~C. Ribeiro, A.~H.~R. Costa, Heuristic selection of
  actions in multiagent reinforcement learning, in: IJCAI 2007, Proceedings of
  the 20th International Joint Conference on Artificial Intelligence,
  Hyderabad, India, January 6-12, 2007, 2007.

\bibitem{Bianchi2014Heuristically}
R.~A. Bianchi, M.~F. Martins, C.~H. Ribeiro, A.~H. Costa,
  Heuristically-accelerated multiagent reinforcement learning, IEEE
  Transactions on Cybernetics 44~(2) (2014) 252.

\bibitem{DBLP:journals/corr/LoweWTHAM17}
R.~Lowe, Y.~Wu, A.~Tamar, J.~Harb, P.~Abbeel, I.~Mordatch,
  \href{http://arxiv.org/abs/1706.02275}{Multi-agent actor-critic for mixed
  cooperative-competitive environments}, CoRR abs/1706.02275.
\newblock \href {http://arxiv.org/abs/1706.02275} {\path{arXiv:1706.02275}}.
\newline\urlprefix\url{http://arxiv.org/abs/1706.02275}

\bibitem{lillicrap2016continuous}
T.~Lillicrap, J.~J. Hunt, A.~Pritzel, N.~Heess, T.~Erez, Y.~Tassa, D.~Silver,
  D.~Wierstra, Continuous control with deep reinforcement learning.

\bibitem{he2016opponent}
H.~He, J.~Boyd-Graber, K.~Kwok, H.~Daum{\'e}~III, Opponent modeling in deep
  reinforcement learning, in: International Conference on Machine Learning,
  2016, pp. 1804--1813.

\bibitem{mnih2015human}
V.~Mnih, K.~Kavukcuoglu, D.~Silver, A.~A. Rusu, J.~Veness, M.~G. Bellemare,
  A.~Graves, M.~Riedmiller, A.~K. Fidjeland, G.~Ostrovski, et~al., Human-level
  control through deep reinforcement learning, Nature 518~(7540) (2015)
  529--533.

\bibitem{wilson1995classifier}
S.~W. Wilson, Classifier fitness based on accuracy, Evolutionary computation
  3~(2) (1995) 149--175.

\bibitem{wilson1998generalization}
S.~W. Wilson, S.~Wilson, G.~Xcs, et~al., Generalization in the xcs classifier
  system.

\bibitem{Holland1992Adaptation}
J.~H. Holland, Adaptation in Natural and Artificial System, 1992.

\bibitem{li2006refined}
J.-B. Li, A.-P. Chen, Refined group learning based on xcs and neural network in
  intelligent financial decision support system, in: Sixth International
  Conference on Intelligent Systems Design and Applications, Vol.~2, IEEE,
  2006, pp. 925--930.

\bibitem{chen2009hierarchical}
K.-Y. Chen, P.~A. Lindsay, P.~J. Robinson, H.~A. Abbass, A hierarchical
  conflict resolution method for multi-agent path planning, in: 2009 IEEE
  Congress on Evolutionary Computation, IEEE, 2009, pp. 1169--1176.

\bibitem{inoue2005exploring}
H.~Inoue, K.~Takadama, K.~Shimohara, Exploring xcs in multiagent environments,
  in: Proceedings of the 7th annual workshop on Genetic and evolutionary
  computation, ACM, 2005, pp. 109--111.

\bibitem{lode2010adaption}
C.~Lode, U.~Richter, H.~Schmeck, Adaption of xcs to multi-learner predator/prey
  scenarios, in: Proceedings of the 12th annual conference on Genetic and
  evolutionary computation, ACM, 2010, pp. 1015--1022.

\bibitem{Silva2006Dealing}
B.~C.~D. Silva, E.~W. Basso, A.~L.~C. Bazzan, P.~M. Engel, Dealing with
  non-stationary environments using context detection, in: Machine Learning,
  Proceedings of the Twenty-Third International Conference (ICML 2006),
  Pittsburgh, Pennsylvania, USA, June 25-29, 2006, 2006.

\bibitem{hernandezleal2017towards}
P.~Hernandezleal, M.~Kaisers, Towards a fast detection of opponents in repeated
  stochastic games (2017) 239--257.

\bibitem{NIPS2018_7374}
Y.~ZHENG, Z.~Meng, J.~Hao, Z.~Zhang, T.~Yang, C.~Fan,
  \href{http://papers.nips.cc/paper/7374-a-deep-bayesian-policy-reuse-approach-against-non-stationary-agents.pdf}{A
  deep bayesian policy reuse approach against non-stationary agents}, in:
  S.~Bengio, H.~Wallach, H.~Larochelle, K.~Grauman, N.~Cesa-Bianchi, R.~Garnett
  (Eds.), Advances in Neural Information Processing Systems 31, Curran
  Associates, Inc., 2018, pp. 954--964.
\newline\urlprefix\url{http://papers.nips.cc/paper/7374-a-deep-bayesian-policy-reuse-approach-against-non-stationary-agents.pdf}

\bibitem{Veloso06probabilisticpolicy}
M.~Veloso, Probabilistic policy reuse in a reinforcement learning agent, in: In
  AAMAS, 2006, pp. 720--727.

\bibitem{piselection}
S.~Li, C.~Zhang, \href{http://arxiv.org/abs/1709.08201}{An optimal online
  method of selecting source policies for reinforcement learning}, CoRR
  abs/1709.08201.
\newblock \href {http://arxiv.org/abs/1709.08201} {\path{arXiv:1709.08201}}.
\newline\urlprefix\url{http://arxiv.org/abs/1709.08201}

\bibitem{zhang2020kogun}
P.~Zhang, J.~Hao, W.~Wang, H.~Tang, Y.~Ma, Y.~Duan, Y.~Zheng, Kogun:
  Accelerating deep reinforcement learning via integrating human suboptimal
  knowledge (2020).
\newblock \href {http://arxiv.org/abs/2002.07418} {\path{arXiv:2002.07418}}.

\bibitem{claus1998dynamics}
C.~Claus, C.~Boutilier, The dynamics of reinforcement learning in cooperative
  multiagent systems, AAAI/IAAI 1998 (1998) 746--752.

\bibitem{weinberg2004best}
M.~Weinberg, J.~S. Rosenschein, Best-response multiagent learning in
  non-stationary environments, in: Proceedings of the Third International Joint
  Conference on Autonomous Agents and Multiagent Systems-Volume 2, IEEE
  Computer Society, 2004, pp. 506--513.

\bibitem{Ganzfried2011Game}
S.~Ganzfried, T.~Sandholm, Game theory-based opponent modeling in large
  imperfect-information games, in: 10th International Conference on Autonomous
  Agents and Multiagent Systems (AAMAS 2011), Taipei, Taiwan, May 2-6, 2011,
  Volume 1-3, 2011.

\bibitem{DPIQN}
Z.~Hong, S.~Su, T.~Shann, Y.~Chang, C.~Lee,
  \href{http://arxiv.org/abs/1712.07893}{A deep policy inference q-network for
  multi-agent systems}, CoRR abs/1712.07893.
\newblock \href {http://arxiv.org/abs/1712.07893} {\path{arXiv:1712.07893}}.
\newline\urlprefix\url{http://arxiv.org/abs/1712.07893}

\bibitem{XOMQ}
H.~Chen, C.~Wang, J.~Huang, J.~Kong, H.~Deng,
  \href{https://doi.org/10.1016/j.neucom.2020.02.118}{Xcs with opponent
  modelling for concurrent reinforcement learners}, Neurocomputing.
\newline\urlprefix\url{https://doi.org/10.1016/j.neucom.2020.02.118}

\bibitem{uther1997adversarial}
W.~Uther, M.~Veloso, Adversarial reinforcement learning, Tech. rep., Technical
  report, Carnegie Mellon University, 1997. Unpublished (1997).

\bibitem{shoham09a}
Y.~Shoham, K.~Leyton-Brown,
  \href{http://jmvidal.cse.sc.edu/library/shoham09a.pdf}{Multiagent Systems:
  Algorithmic, Game-Theoretic, and Logical Foundations}, Cambridege University
  Press, 2009.
\newline\urlprefix\url{http://jmvidal.cse.sc.edu/library/shoham09a.pdf}

\bibitem{hernandezleal2014a}
P.~Hernandezleal, E.~M. De~Cote, L.~E. Sucar, A framework for learning and
  planning against switching strategies in repeated games, adaptive and
  learning agents 26~(2) (2014) 103--122.

\bibitem{mnih2016asynchronous}
V.~Mnih, A.~P. Badia, M.~Mirza, A.~Graves, T.~Lillicrap, T.~Harley, D.~Silver,
  K.~Kavukcuoglu, Asynchronous methods for deep reinforcement learning, in:
  International conference on machine learning, 2016, pp. 1928--1937.

\bibitem{heess2017emergence}
N.~Heess, T.~B. Dhruva, S.~Sriram, J.~Lemmon, J.~Merel, G.~Wayne, Y.~Tassa,
  T.~Erez, Z.~Wang, S.~M.~A. Eslami, et~al., Emergence of locomotion behaviours
  in rich environments, arXiv: Artificial Intelligence.

\bibitem{schulman2017proximal}
J.~Schulman, F.~Wolski, P.~Dhariwal, A.~Radford, O.~Klimov, Proximal policy
  optimization algorithms, arXiv preprint arXiv:1707.06347.

\bibitem{lowe2017multi}
R.~Lowe, Y.~Wu, A.~Tamar, J.~Harb, O.~P. Abbeel, I.~Mordatch, Multi-agent
  actor-critic for mixed cooperative-competitive environments, in: Advances in
  Neural Information Processing Systems, 2017, pp. 6379--6390.

\bibitem{raileanu2018modeling}
R.~Raileanu, E.~Denton, A.~Szlam, R.~Fergus, Modeling others using oneself in
  multi-agent reinforcement learning, in: J.~Dy, A.~Krause (Eds.), Proceedings
  of the 35th International Conference on Machine Learning, Vol.~80 of
  Proceedings of Machine Learning Research, PMLR, Stockholmsmässan, Stockholm
  Sweden, 2018, pp. 4257--4266.

\bibitem{hernandezleal2016identifying}
P.~Hernandezleal, M.~E. Taylor, B.~Rosman, L.~E. Sucar, E.~M. De~Cote,
  Identifying and tracking switching, non-stationary opponents: A bayesian
  approach.

\bibitem{rosman2016bayesian}
B.~Rosman, M.~Hawasly, S.~Ramamoorthy, Bayesian policy reuse, Machine Learning
  104~(1) (2016) 99--127.

\bibitem{yang2019towards}
T.~Yang, J.~Hao, Z.~Meng, C.~Zhang, Y.~Zheng, Z.~Zheng, Towards efficient
  detection and optimal response against sophisticated opponents (2019)
  623--629.

\bibitem{van2011insights}
H.~P. van Hasselt, Insights in reinforcement learning, Hado van Hasselt, 2011.

\bibitem{LiuMultiobjective}
C.~Liu, X.~Xu, D.~Hu, Multiobjective reinforcement learning: A comprehensive
  overview, IEEE Trans Cybern 45~(3)  385--398.

\end{thebibliography}

\begin{appendices}
	
	\section{Parameters of the algorithms used in the experiments}\label{appendix}
	\subsection{Initial parameters of HAMXCS in the experiments}
	\begin{table}[h]\label{initial hamxcs}
		\centering
		\caption{Initial parameters of HAMXCS in the experiments.}
		\label{Initial parameters} 
		\begin{tabular}{lll}	
			\hline\noalign{\smallskip}
			Parameter & Notation & value  \\
			\noalign{\smallskip}\hline\noalign{\smallskip}
			Initial prediction vector&$ \bm{p} $&[0.00001 0.00001 $ \cdots $]\\
			Initial heuristic vector&$ \bm{h} $&[0.00001 0.00001 $ \cdots $]\\
			Initial prediction error&$ \epsilon $&0.00001\\
			Initial fitness&$ F $&0.00001\\
			Population size &$ N $ & 500\\
			Heuristic update magnitude &$\rho$&10\\
			Heuristic weight &$\kappa$&1\\
			Entropy parameter &$\eta$&0.001\\
			Trace decay & $\lambda$ &0.05\\
			Learning rate & $\beta_i$ $ (i=1,2,3,4,5 )$ & 0.15 \\
			Accuracy coefficient & $\alpha$ & 0.1 \\
			Error threshold & $ \epsilon_{0} $ & 0.01\\
			Accuracy power & $ \nu $ & 5\\
			Discount factor & $\gamma$ & 0.71\\
			GA threshold & $ \theta_{GA} $ & 35\\
			Eligibility trace threshold & $ \theta_{et} $&0.001\\
			Cross probability & $\chi$ & 0.75\\
			Mutation probability &$\mu$ &0.03\\
			Deletion threshold &$ \theta_{del} $&20\\
			Fitness threshod & $\delta$&0.1\\
			Subsumption threshold&$ \theta_{sub} $&20\\
			Wildcard probability&$ P_{\#} $&0.33\\
			\noalign{\smallskip}\hline
		\end{tabular}
	\end{table}
	
	\subsection{Parameters of A3C in the experiments}
	\begin{table}[h]
		\label{initial A3C}
		\centering
		\caption{Parameters of A3C in the experiments.}
		\label{A3C params} 
		\begin{tabular}{|c|c|c|c|}	
			\hline
			Parallel learners & 4 &Discount factor & 0.9  \\
			\hline
			Actor learning rate & 0.0001 &Critic learning rate &0.0001\\
			\hline
			Entropy parameter & 0.001 &Optimizer &RMSPropOptimizer\\
			\hline
			Activation function & ReLU &  & \\
			\hline
		\end{tabular}
	\end{table}
	Parameters of A3C used in the experiments are shown in Table \ref{A3C params}. In Hexcer, the A3C agent was updated every 5 matches, or when the agent scored a goal. In thief-and-hunter scenario, the A3C agent was updated every 5 matches, or when it got the goal position. 
	
	\subsection{Parameters of MADDPG in the experiments}
	\begin{table}[h]\label{initial MADDPG}
		\centering
		\caption{Parameters of MADDPG in the experiments.}
		\label{MADDPG params} 
		\resizebox{\linewidth}{!}
		{
			\begin{tabular}{|c|c|c|c|}	
				\hline
				Memory size & 1000 & Minibatch size & 100 \\
				\hline
				Actor learning rate & 0.0001 &Critic learning rate &0.002 (Hexcer) / 0.001 (Thief-and-hunter)\\
				\hline
				Discount factor & 0.9 &Optimizer &AdamOptimizer\\
				\hline
				Activation function & ReLU &Soft replacement parameter &0.01\\
				\hline
				Learning rate for the estimated player model & 0.0001 &Entropy parameter & 0.001\\
				\hline
				Update frequency & every 50 steps & & \\
				\hline
			\end{tabular}
		}
	\end{table}
	Parameters of MADDPG used in the experiments are shown in Table \ref{MADDPG params}. The output layer of the actor in MADDPG was a softmax layer with each unit corresponding to the candidate action selection probability. Besides, the estimated player model had a hidden layer of 100 units. The MADDPG agent began to updated after 1000 transactions stored in the memory.

	\subsection{Parameters of DQN in the experiments}
	\begin{table}[h]\label{initial DQN}
		\centering
		\caption{Parameters of DQN in the experiments.}
		\label{DQN params} 
		\resizebox{\linewidth}{!}
		{
			\begin{tabular}{|c|c|c|c|}	
				\hline
				Learning rate & 0.001 (Hexcer) / 0.0001 (Thief-and-hunter) & Minibatch size & 32 \\
				\hline
				Optimizer & RMSPropOptimizer & Discount factor & 0.9 \\
				\hline
				Memory size & 10000 & Target network & Update every 5 matches \\
				\hline
				$\epsilon-greedy$ & 1.0(initial)$\rightarrow$0.001 &  &  \\
				\hline
			\end{tabular}
		}
	\end{table}
	Parameters of DQN used in the experiments are shown in Table \ref{DQN params}. In the experiments, the DQN agent began to updated after 500 transactions stored in the memory.
	
	\subsection{Parameters of PPO in the experiments}
	\begin{table}[h]\label{initial PPO}
		\centering
		\caption{Parameters of PPO in the experiments.}
		\label{PPO params} 
		\resizebox{\linewidth}{!}
		{
			\begin{tabular}{|c|c|c|c|}	
				\hline
				Discount factor & 0.9 & Minibatch size & 32 \\
				\hline
				Actor learning rate & 0.0001 &Critic learning rate &0.0001 \\
				\hline
				Clipping parameter & 0.2 &Optimizer &AdamOptimizer\\
				\hline
				Activation function & ReLU &Soft replacement parameter &0.01\\
				\hline
				Actor update rounds & 10 &Critic update rounds &10 \\
				\hline
				Worker &1& &\\
				\hline
				
			\end{tabular}
		}
	\end{table}
	Parameters of PPO used in the experiments are shown in Table \ref{PPO params}. In Hexcer, the PPO agent was updated every 5 matches, or when the agent scored a goal. 
	
	\subsection{Parameters of DPPO in the experiments}
	\begin{table}[h]\label{initial DPPO}
		\centering
		\caption{Parameters of DPPO in the experiments.}
		\label{DPPO params} 
		\resizebox{\linewidth}{!}
		{
			\begin{tabular}{|c|c|c|c|}	
				\hline
				Parallel workers & 4 & Minibatch size & 32 (Hexcer) / 100 (Thief-and-hunter)\\
				\hline
				Actor learning rate & 0.0001 (Hexcer) / 0.00005 (Thief-and-hunter) &Critic learning rate &0.0001 (Hexcer) / 0.0005 (Thief-and-hunter)\\
				\hline
				Clipping parameter & 0.2 &Optimizer &AdamOptimizer\\
				\hline
				Activation function & ReLU &Soft replacement parameter &0.01\\
				\hline
				Actor update rounds & 10 &Critic update rounds &10\\
				\hline
				Discount factor & 0.9 & & \\
				\hline
			\end{tabular}
		}
	\end{table}
	Parameters of DPPO used in the experiments are shown in Table \ref{DPPO params}. In Hexcer, the DPPO agent was updated every 5 matches, or when the agent scored a goal. In thief-and-hunter scenario, the DPPO agent was updated every 100 steps, or when it got the goal position. Besides, in DPPO, we used the clipped surrogate technique in PPO.
\end{appendices}
\end{document}